\documentclass[11pt]{idea_style}
\newif\ifdraft
\draftfalse

\usepackage[T1]{fontenc}
\usepackage{url}
\usepackage{nicefrac}
\usepackage{multicol}
\usepackage{microtype}
\usepackage{geometry}
\usepackage{graphicx}%
\graphicspath{{fig/}}
\usepackage{multirow}%
\usepackage{amsmath}
\usepackage{amssymb}
\usepackage{amsfonts}
\usepackage{stmaryrd}
\usepackage{mathrsfs}%
\usepackage{xcolor}%
\usepackage{textcomp}%
\usepackage{manyfoot}%
\usepackage{booktabs}%
\usepackage{algorithm}%
\usepackage{tabularx}%
\usepackage{algorithmicx}%
\usepackage{algpseudocode}%
\usepackage{listings}%
\usepackage[inline]{enumitem}
\usepackage{xspace}
\usepackage{tikz}
\usepackage{makecell}
\usepackage{longtable}
\usepackage{changepage}
\usepackage{color}
\usepackage{colortbl}
\usepackage{pifont}
\usepackage{subcaption}
\usepackage{physics}
\usepackage{diagbox}
\usepackage{wrapfig}
\usepackage{siunitx}
\usepackage[misc]{ifsym} 
\usepackage[numbers]{natbib}
\usepackage{cuted}
\usepackage{capt-of}
\usepackage{hyperref}
\hypersetup{
  colorlinks,
  linkcolor=ideacolor,
  citecolor=ideacolor,
  urlcolor=ideacolor
}

\usepackage{setspace}
\usepackage{changepage}
\usepackage{float}
\usepackage{tablefootnote}
\usepackage{threeparttable}
\usepackage{placeins}

\usepackage{amsmath,amsfonts,bm}

\def\eqref#1{equation~\ref{#1}}
\def\1{\bm{1}}

\DeclareMathAlphabet{\mathsfit}{\encodingdefault}{\sfdefault}{m}{sl}
\SetMathAlphabet{\mathsfit}{bold}{\encodingdefault}{\sfdefault}{bx}{n}

\definecolor{myorange}{RGB}{220,128,0}
\definecolor{myblue}{RGB}{0,80,220}
\definecolor{myred}{RGB}{220,20,0}

\newcommand{\ie}{\emph{i.e.},\xspace}
\newcommand{\eg}{\emph{e.g.},\xspace}

\ifdraft
    \providecommand\todo[1]{[\textcolor{red}{TODO: {#1}}]}
\else
    \providecommand\todo[1]{}
\fi

\author[]{%
    Youqiang Gui$^{1,2,*}$,
    Yunfei Liu$^{2,\dagger,\ddagger}$,
    Lijian Lin$^{2}$,
    Ye Zhu$^{2}$,
    Jianan Wang$^{3}$,
    Cheng Peng$^{1,\ddagger}$,
    Yu Li$^{2}$
}

\affiliation{%
    $^{1}$ Sichuan University
    $^{2}$ International Digital Economy Academy (IDEA)
    $^{3}$ Astribot
}

\ideadata[$^{*}$]{This work was done during an internship at IDEA.}
\ideadata[$^{\dagger}$]{Project leader.}
\ideadata[$^{\ddagger}$]{Corresponding authors.}

\ideadata[Project page]{\url{https://youqiang-gui.github.io/EIP/}}

\title{Embodied Interaction Policy: 

A Unified Model for Full-Duplex Human--Robot Interaction}

\abstract{
\small{
\begin{abstract}

Real-time human-robot social interaction requires continuous coordination between speech and physical behavior. Existing systems often separate interaction agents from motion generation, complicating coordination as conversational states change. We propose \textbf{Embodied Interaction Policy (EIP)}, a unified streaming framework that couples an Audio-Interaction expert with a motion expert through layer-wise joint attention. This allows motion generation to directly access the evolving interaction context. A chunk-wise causal mask prevents access to future context, while a vocal token encodes speaking state and the semantic emotion of the response. We construct \textbf{Embodied Interaction Dataset (EI-Data)} from co-speech and dyadic interaction data and train our model in two stages to learn state-adaptive and emotion-conditioned behavior.
Combined with the frozen interaction expert, EIP supports continuous
robot motion during proactive responses and interruption-aware turn-taking. Quantitative evaluations demonstrate improved motion quality over representative gesture-generation baselines. Compared with a representative cascaded system, EIP achieves higher interaction success and reduces warm-start latency by 64\%. Physical deployment and user studies further demonstrate improved motion naturalness and interaction comfort.
\end{abstract}
}
}

\begin{document}

\maketitle

\begin{figure}[h]
    \centering
    \includegraphics[width=\linewidth]{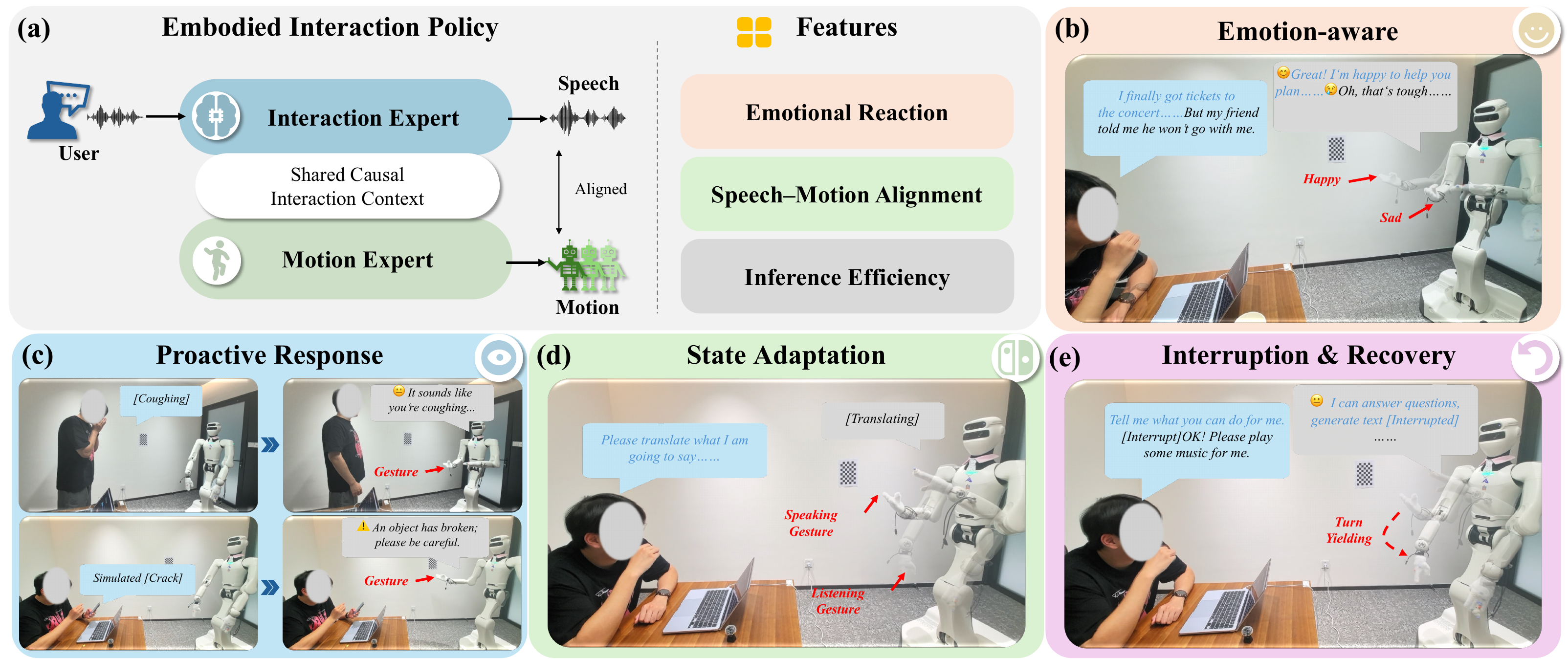}

    \captionsetup{hypcap=false}
    \captionof{figure}{
        (a) We propose Embodied Interaction Policy (EIP),
        which is a unified streaming architecture that
        couples interaction and motion experts for
        speech--motion alignment.
        EIP enables diverse interaction scenarios,
        including (b) emotion-aware gesture generation,
        (c) proactive response,
        (d) seamless state adaptation, and
        (e) fast reaction to user interruption
        and task recovery.
    }
    \label{fig:abstract}
    \captionsetup{hypcap=true}
\end{figure}

\section{Introduction}

\begin{figure*}[t]
    \centering
    \includegraphics[width= 1\linewidth]{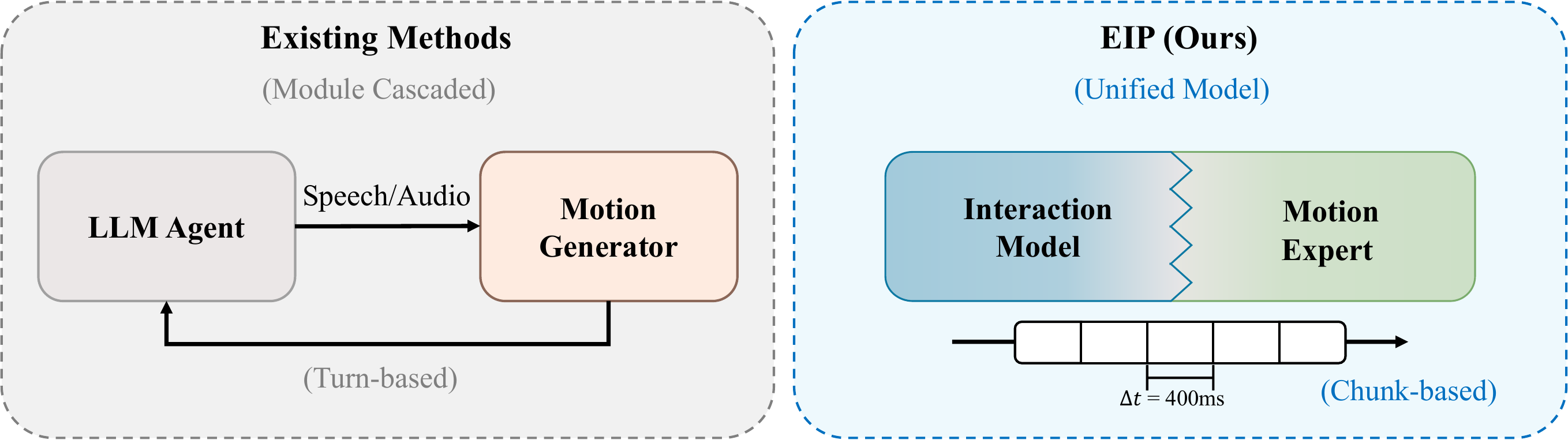}
    
    \caption{\textbf{Comparison of embodied interaction modeling paradigms. }Existing  embodied interaction methods typically follow a cascaded, turn-based paradigm. In contrast, our approach unifies the interaction model and motion expert within a single architecture and performs continuous chunk-based processing at a 400-ms temporal granularity.
}
    \label{fig:comparison}
\end{figure*}

% Our contributions are threefold:
% \begin{itemize}
%     \item We present the first unified streaming
%     framework for real-time embodied human--robot interaction that jointly
%     generates speech and continuous motion.
%     \item We introduce EIP, a unified streaming framework that supports proactive embodied responses to salient acoustic events, interruption-aware turn-taking, and gestures that adapt to the robot's interaction state and response emotion.
%     \item We construct EI-Bench for training and evaluating embodied interaction, and deploy EIP on the Astribot S1. Quantitative experiments and user studies validate motion quality, interaction success, response latency, and user experience.
% \end{itemize}

Natural human–robot social interaction requires a robot to continuously determine what to say, when to speak, and how to move during interaction~\citep{zhang2026proact, wang2026robogesture}. Beyond generating appropriate verbal responses, the robot must exhibit natural and smoothly transitioning physical behaviors that remain consistent with the evolving conversational context. In particular, the robot should maintain smooth and expressive motion as it switches between interaction states, such as listening, speaking, and responding to interruptions~\citep{skantze2021turntaking}. Such nonverbal behavior is closely coupled with conversational timing: decisions about when to respond, remain silent, or yield to an interruption directly affect how physical movements should initiate, sustain, and transition. Achieving this tight temporal coordination between interaction and motion remains a key challenge in embodied social robotics.

Existing co-speech gesture generation methods have shown great success in generating realistic and diverse body, hand, and facial motions aligned with speech \citep{yi2023generating,liu2024emage,mughal2026miburi}. However, these methods typically generate motion in human-centric representation and in digital space (\eg, SMPL-X~\citep{pavlakos2019expressive}). Using retargeting postprocessing method~\cite{araujo2026retargeting} can be a solution for adapting to robot. However, such retargeting may introduce kinematic discrepancies that compromise its expressiveness and temporal fidelity. To address this embodiment gap, recent approaches propose to generate robot motion for physical deployment. ProAct~\citep{zhang2026proact} combines proactive reasoning with continuous robot motion generation, while retaining turn-based verbal responses. RoboGesture~\citep{wang2026robogesture} integrates an Automatic Speech Recognition (ASR)--Large Language Model (LLM)--Text-to-Speech (TTS) pipeline with a streaming speech-to-motion generator for humanoid deployment. 

Although these systems enable embodied interaction to some extent, they largely follow a turn-based interaction paradigm, where the robot waits for a user turn to complete before generating a verbal response and the corresponding motion. This design couples interaction responsiveness with the latency of the upstream ASR--LLM--TTS pipeline, limiting rapid responses to user utterances and interruptions (Fig.~\ref{fig:comparison}, left). More recent duplex interaction models, such as Audio Interaction Model (AIM) \citep{xie2026audio}, remove this constraint by continuously processing the ongoing audio stream and supporting response generation without waiting for a complete turn to finish. However, timely interaction responses alone do not guarantee natural physical behavior. As the interaction state changes continuously, the robot must also smoothly initiate, sustain, and transition its physical behavior accordingly, rather than switching abruptly between discrete motion states. How to translate such evolving interaction context into continuous, expressive, and temporally coherent robot motion in real time remains unexplored.

Our key observation is that the internal representations of a pretrained interaction model already encode rich, continuously evolving cues about the ongoing interaction, and can therefore serve as a natural context for robot motion generation. Building on this insight, we propose \textbf{Embodied Interaction Policy (EIP)}, a unified streaming framework that grounds such interaction representations into continuous physical motion. EIP introduces a dedicated motion expert that directly consumes the evolving interaction context through layer-wise interaction–motion coupling, allowing robot motion to be generated in lockstep with conversational processing. To preserve causality during streaming generation, we introduce a chunk-wise causal attention mask that restricts motion generation to interaction information available up to the current time step. Finally, we introduce a vocal token that encodes the robot’s speaking state and response emotion, enabling more state-appropriate and expressive motion. Furthermore, to learn such policies and address the lack
of systematic evaluation for continuous embodied interaction,
we construct the \textbf{Embodied Interaction Dataset (EI-Data)} from existing co-speech and dyadic interaction datasets. Our contributions are threefold:

\begin{figure*}[t]
    \centering
    \includegraphics[width= 1\linewidth]{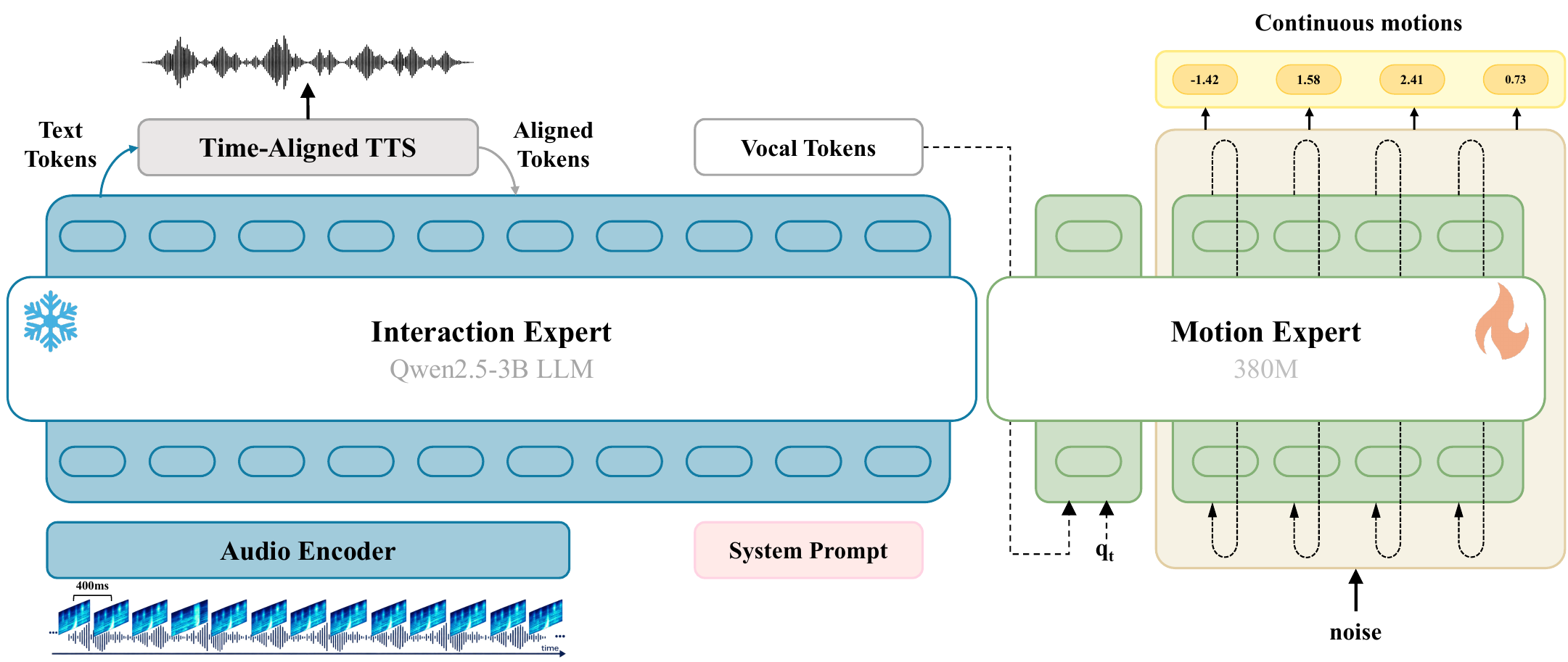}
    \caption{\textbf{Overview of the EIP framework.} EIP consists of a pretrained interaction expert that processes streaming audio and models the evolving interaction context, together with a motion expert that generates continuous and synchronized robot motions.
}
    \label{fig:pipeline}
    \vspace{-4pt}
\end{figure*}

\begin{itemize}
    \item  We reveal that the hidden states of interaction experts can effectively serve as interaction context for robot motion generation. Building on this insight, we develop \textbf{EIP}, a unified streaming framework that couples interaction and motion experts through layer-wise attention to coordinate streaming speech and continuous robot motion.
    \item We introduce a chunk-wise causal mask to preserve causality in
    streaming motion generation, together with a vocal token for
    state- and emotion-aware motion control.
    \item We construct \textbf{EI-Data} for training and evaluating embodied interaction, and deploy EIP on the Astribot S1. Quantitative experiments and user studies validate motion quality, interaction success, response latency, and user experience.
\end{itemize}

\section{Related Work}
\label{sec:Realted Work}

\subsection{Human--Robot Interaction}

Social interaction has been increasingly explored through virtual humans and autonomous characters, where multimodal social behaviors can be modeled without the physical constraints of robotic embodiment \citep{cai2024digital,jiang2025solami}. 
Conventional approaches often learn motion in human representation and subsequently retarget it to the robot~\citep{yoon2019robots}. Recent retargeting methods improve the fidelity of this transfer \citep{araujo2026retargeting,yang2026omniretarget}. 
However, transferring such behaviors to physical robots introduces substantial embodiment gaps, including morphological differences, kinematic feasibility, safety, expressiveness degradation, and real-time execution.

Recent robot-centric work has therefore shifted toward learning directly in the robot's representation, moving cross-embodiment alignment from inference-time post-processing to data preprocessing~\citep{zhang2026proact,wang2026robogesture}. 
Building on this direction, ProAct~\citep{zhang2026proact} combines a streaming multimodal interaction model with a streaming motion generator for real-time embodied social interaction. RoboGesture~\citep{wang2026robogesture} learns streaming co-speech motion directly in robot space and deploys it on a physical humanoid robot. 
Despite these advances, verbal interaction and motion generation remain
cascaded, introducing additional inter-module handoff and synchronization overhead that can increase end-to-end latency. 
In contrast, EIP unifies interaction and motion generation within a unified policy, enabling low-latency and synchronized embodied interaction.

\subsection{Interaction Models}

Interactive agents have evolved from text-based systems that interact through language and tools~\citep{yao2023react,schick2023toolformer}, to speech-based full-duplex models~\citep{defossez2024moshi,ma2025language}, and multimodal models that continuously process audio and visual inputs~\citep{xu2025qwen25omni}.
Recent work further treats interaction itself as a native modeling problem.
Thinking Machines' Interaction Model~\citep{thinkingmachines2026interactionmodels} models audio, video, and text as time-aligned continuous streams, enabling interruption, overlap, and responsive interaction.
Audio Interaction Model ~\citep{xie2026audio} formulates a similar paradigm for continuous audio through an always-on perceive--decide--respond loop. These works primarily model interaction through digital modalities such as
speech, text, and vision. We extend this interaction paradigm to physical
embodiment, incorporating robot motion into the continuous interaction loop.

\subsection{Co-Speech Gesture Generation}
Another related topic is co-speech gesture generation, which generates expressive motion in sync with speech. Recent methods have advanced from holistic speech-driven motion generation~\citep{yi2023generating,liu2024emage} to photorealistic conversational motion~\citep{ng2024photoreal} and causal online gesture generation~\citep{mughal2026miburi}. 
While these methods primarily generate digital motion conditioned on the speaker's speech, EIP targets continuous embodied dialogue interaction, jointly modeling robot behaviors across speaking, listening, and idle states.

\section{Embodied Interaction Policy}

\subsection{Overview}

As illustrated in Fig.~\ref{fig:pipeline}, EIP is a unified streaming policy that couples a frozen interaction expert with a compact motion expert. 
The interaction model continuously processes human audio, provides rich context, and determines whether the robot should respond and how the robot responds, while the motion expert generates the corresponding robot behavior. 
Following~\citep{xie2026audio}, both experts operate on a shared interaction clock of $\Delta=400$~ms. 
At interaction step $t$, EIP takes the latest human-audio chunk $\mathbf{x}_t$, the robot state $\mathbf{q}_t$,
and the accumulated context $\mathbf{h}_{t-1}$, and produces synchronized speech and motion chunks for the next interval:
\begin{equation}
(\hat{\mathbf{s}}_{t},\hat{\mathbf{a}}_{t})
=
\pi_{\mathrm{EIP}}
(\mathbf{x}_t,\mathbf{q}_t, \mathbf{h}_{t-1}).
\label{eq:eip_policy}
\end{equation}
Here, $\hat{\mathbf{s}}_{t} \in \mathbb{R}^{\Delta \cdot r_s}$ denotes a 400-ms speech chunk,
and $\hat{\mathbf{a}}_{t} \in \mathbb{R}^{\Delta \cdot r_a \times N}$ denotes the robot motion chunk. Here $r_s, r_a, N$ are the sample rate of audio, the motion frequency, and the dimension of action space, respectively.

EIP is a unified model through layer-wise joint attention between the interaction and motion experts, allowing motion generation to access the evolving interaction context directly. 
To preserve causality, a chunk-wise attention mask blocks future interaction context and motion information from other chunks. We further introduce a vocal token that explicitly encodes the robot's speaking state and the semantic emotion of its response, thereby aligning physical behavior with verbal output. Next, we elaborate on the design and implementation of our method.

% ================================================================
% 3.2 Unified Interaction--Motion Architecture
% ================================================================

\subsection{Unified Interaction--Motion Architecture}
\label{sec:unified-architecture}

\paragraph{Audio interaction expert.}
We initialize the interaction expert with the pre-trained Audio-Interaction model~\citep{xie2026audio}, which follows the perceive--decide--respond paradigm. 
Given the current human-audio chunk $\mathbf{x}_t$, it updates the interaction context and predicts whether and how to respond:
\begin{equation}
d_t = F_{\mathrm{dec}}(\mathbf{h}_t),
\qquad \text{here}~~
\mathbf{h}_t = F_{\mathrm{per}} \bigl( \mathbf{h}_{t-1}, E_{\mathrm{aud}}(\mathbf{x}_t) \bigr),
\end{equation}
where $d_t\in\{\langle\mathrm{silent}\rangle,
\langle\mathrm{response}\rangle\}$ denotes the silence--response decision, $E_{\mathrm{aud}}$ is the audio encoder,
$F_{\mathrm{per}}$ updates the interaction context using the current
audio input, and $F_{\mathrm{dec}}$ maps the updated context $h_t$
to a corresponding silence--response decision.
When $d_t=\langle\mathrm{silent}\rangle$, the output is silent, \ie, $ \hat{\mathbf{s}}_t=\varnothing$; otherwise, the response module $F_{\mathrm{resp}}$ generates the corresponding response audio.
\begin{equation}
\hat{\mathbf{s}}_{t} =
\begin{cases}
\varnothing, & d_t=\langle\mathrm{silent}\rangle,\\
F_{\mathrm{resp}}(\mathbf{h}_t), & d_t=\langle\mathrm{response}\rangle.
\end{cases}
\end{equation}

\paragraph{Motion expert.}
Inspired by the design of $\pi_0$\citep{black2024pi_0}, we design a flow-matching based expert for motion generation. For each interaction step $t$, the motion expert receives the current robot state $\mathbf{q}_t$, a vocal condition $\mathbf{v}_t$ (which will be introduced in Eq. (\ref{eq:vocal-token})), and a noisy motion trajectory $\mathbf{a}^{\tau}_t$. 
First, the robot state and noisy motion are projected into the motion-expert hidden space as
\begin{equation}
\mathbf{Q}_t = \mathcal{E}_{\mathrm{state}}(\mathbf{q}_t),~\mathbf{A}^{\tau}_t =
\mathcal{E}_{\mathrm{act}}
\left(
    \mathbf{a}^{\tau}_t,
    \mathcal{E}_{\tau}(\tau)
\right).
\end{equation} \label{eq:motion-tokenization}
Here, $\mathbf{q}_t, \mathbf{a}^{\tau}_t \in \mathbb{R}^{(\Delta r_a)\times N}$ are the robot-state
sequence and noisy motion trajectory at step $t$, respectively.
$\mathcal{E}_{\mathrm{state}}$ is a state encoder, and position encoder $\mathcal{E}_{\tau}$ encodes the flow time
$\tau$, which is combined with the noisy motion by a fusion block $\mathcal{E}_{\mathrm{act}}$. The resulting state
and motion token block $\mathbf{Q}_t,\mathbf{A}^{\tau}_t\in\mathbb{R}^{(\Delta r_a)\times d_m}$, where $d_m$ is the token
dimension, form the motion chunk together with the vocal token
$\mathbf{v}_t$:
\begin{equation}
\mathbf{Z}^{(0)}_t
=
\left[
    \mathbf{Q}_t;
    \mathbf{v}_t;
    \mathbf{A}^{\tau}_t
\right],
\label{eq:motion-suffix}
\end{equation}
where $[\cdots]$ denotes the concatenation. Based on $\mathbf{Z}^{(0)}_t$ and context $\mathbf{h}_{t}$, we generate motion by integrating the learned motion expert net $\mathbf{v}_\theta$ from $\tau = 0$ to $\tau = 1$, starting with random noise $\mathbf{a}^{0}_t \sim \mathcal{N}(\mathbf{0}, \mathbf{I})$. We use forward Euler integration with step size
$\delta=1/K$ for $K$ integration steps:

\begin{equation}
    \mathbf{a}^{\tau + \delta}_t = \mathbf{a}^{\tau}_t + \delta \mathbf{v}_\theta (\mathbf{Z}^{(0)}_t, \mathbf{h}_{t}).
\end{equation}
\paragraph{Layer-wise interaction--motion coupling.}

\begin{wrapfigure}{r}{0.45\columnwidth}
    \centering
    \vspace{-10pt}
    \includegraphics[width=\linewidth]{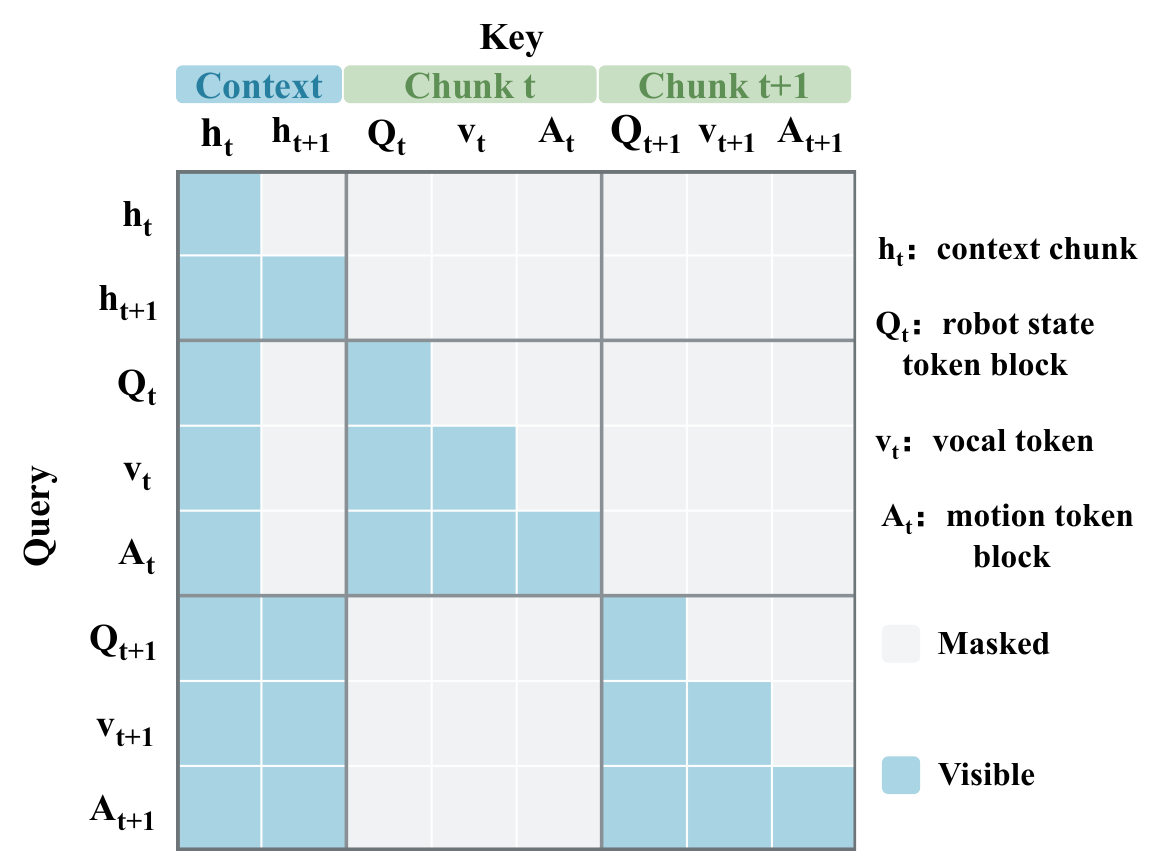}
    \caption{
    \textbf{Chunk-wise causal interaction--motion attention.}
    Blue entries indicate visible attention, whereas gray entries are masked. Interaction context follows causal attention. Each motion chunk accesses interaction context only up to its current time step and follows the within-chunk dependency $\mathbf{Q}_t\rightarrow\mathbf{v}_t\rightarrow\mathbf{A}_t$, with tokens within each block remaining mutually visible. Cross-chunk motion attention is masked.
    }
    \label{fig:mask}
    \vspace{-10pt}
\end{wrapfigure}

Following the Mixture-of-Transformers (MoT) architecture
\citep{zhou2025transfusion, black2024pi_0}, EIP couples a frozen interaction expert with a trainable motion expert through
layer-wise joint attention. To preserve chunk-wise causality, we employ the structured attention mask
illustrated in Fig.~\ref{fig:mask}. Interaction context tokens follow standard causal
attention, while the state tokens $\mathbf{Q}_t$, vocal token $\mathbf{v}_t$,
and motion tokens $\mathbf{A}_t$ can only access the interaction context
available up to chunk $t$. Within each chunk, attention follows
$\mathbf{Q}_t\rightarrow\mathbf{v}_t\rightarrow\mathbf{A}_t$, with tokens
within the same block remaining mutually visible. Cross-chunk motion attention
and access to future interaction context are masked. We justify the token ordering and attention visibility in Appendix~\ref{app:mask_rationale}.

\paragraph{Vocal token conditioning.}
The vocal token combines a speaking-state token with an emotion token. When speaking, the interaction expert
provides an emotion label $m_t$ from six categories:
\begin{equation}
\begin{aligned}
m_t \in \{&
\texttt{happy}, \texttt{sad}, \texttt{angry},\\
&
\texttt{surprise}, \texttt{normal}, \texttt{urgent}\}.
\end{aligned}
\label{eq:emotion-condition}
\end{equation}
When $d_t=\langle\mathrm{silent}\rangle$, we set
$m_t=\texttt{none}$.
The vocal token is then constructed as
\begin{equation}
\mathbf{v}_t
=
\mathcal{E}_{\mathrm{spk}}(d_t)
+
\mathcal{E}_{\mathrm{emo}}(m_t),
\label{eq:vocal-token}
\end{equation}
where $\mathcal{E}_{\mathrm{spk}}$ and $\mathcal{E}_{\mathrm{emo}}$
denote the speaking-state and emotion embeddings, respectively.
The emotion condition $m_t$ additionally modulates the motion-token
features through FiLM, conditioning motion generation on the
semantic emotion of the response.

\subsection{Two-Stage Training}
Jointly learning state adaptation and emotion-aware motion
can lead to mode collapse. Therefore, we train EIP in two stages. Stage I  first learns motion patterns across speaking, listening, and idle states, using only the speaking-state token of the vocal condition. Stage II then introduces emotion-conditioned motion generation by conditioning the motion expert on the full vocal token ($\mathbf{v}_t$) defined in Eq. (\ref{eq:vocal-token}).

\paragraph{Stage I: State adaptation motion learning.}
We train the motion expert on the \textbf{EI-Data} training split
to learn continuous robot motion across speaking, listening and idle
states. The interaction expert remains frozen throughout training.

For a ground-truth motion chunk $\mathbf{a}_t$, we sample Gaussian noise
$\boldsymbol{\epsilon}_t$ and a flow timestep $\tau\in[0,1]$, and construct
the interpolated motion:
\begin{equation}
\mathbf{a}^{\tau}_t
=
(1-\tau)\boldsymbol{\epsilon}_t
+
\tau\mathbf{a}_t,
\qquad
\tau\sim\mathcal{U}(0,1),
\quad
\boldsymbol{\epsilon}_t\sim\mathcal{N}(\mathbf{0},\mathbf{I}),
\label{eq:flow-path}
\end{equation}
where $\mathbf{a}_t$ is the ground-truth motion chunk and
$\mathbf{a}^{\tau}_t$ is its noisy interpolation at flow timestep $\tau$.
For each behavioral phase $r \in \{\mathrm{spk}, \mathrm{lis}, \mathrm{idle}\}$, we optimize
\begin{equation}
\mathcal{L}_{r}
=
\mathbb{E}
\left[
\frac{1}{|\mathcal{M}_{r}|}
\sum_{t\in\mathcal{M}_{r}}
\left\|
\mathbf{v}_{\theta}
\left(
\mathbf{a}^{\tau}_t,
\mathbf{q}_t,
\tau
\mid
\mathbf{h}_t,
\mathbf{v}_t
\right)
-
\left(
\mathbf{a}_t-\boldsymbol{\epsilon}_t
\right)
\right\|_2^2
\right].
\label{eq:phase-flow-matching}
\end{equation}

where $\mathcal{M}_{r}$ selects the motion chunks in phase $r$.
We combine three losses: the co-speech speaking loss
$\mathcal{L}_{\mathrm{spk}}$ learns speech-correlated gestures; the Seamless
listening loss $\mathcal{L}_{\mathrm{lis}}$ exposes the model to diverse
non-speaking reactions while the user is speaking; and the co-speech idle
loss $\mathcal{L}_{\mathrm{idle}}$ encourages stable motion when neither
participant is speaking. The Stage-I objective is:
\begin{equation}
\mathcal{L}_{\mathrm{I}}
=
\lambda_{1}\mathcal{L}_{\mathrm{spk}}
+
\lambda_{2}\mathcal{L}_{\mathrm{lis}}
+
\lambda_{3}\mathcal{L}_{\mathrm{idle}}.
\label{eq:stage-one-objective}
\end{equation}

\paragraph{Stage II: Emotion-aware motion learning.}
Starting from the Stage-I checkpoint, FiLM conditioning is introduced into the final four motion-expert blocks, while the final eight motion-expert blocks and the motion head are jointly fine-tuned.
We sample co-speech examples uniformly across the six emotion categories defined in Eq.~(\ref{eq:emotion-condition}), conditioning each speaking motion chunk on the semantic emotion label $m_t$ of its target response.
We optimize the flow-matching objective over the speaking motion chunks:
\begin{equation}
\mathcal{L}_{\mathrm{II}}
=
\mathbb{E}_{m_t}
\left[
\mathcal{L}_{\mathrm{spk}}(\theta_{\mathrm{II}}; m_t)
\right],
\end{equation}
where $\theta_{\mathrm{II}}$ denotes the FiLM modules and the selected 
motion-expert parameters adapted in Stage II.
During inference, the emotion label predicted by the interaction expert is used as $m_t$ for vocal-token construction and FiLM modulation, enabling motion generation conditioned on the semantic emotion of the response.

% ================================================================
% 3.6 Playback-Synchronized Streaming Inference
% ================================================================

\subsection{Playback-Synchronized Streaming Inference}
\label{sec:streaming_inference}

During training, response tokens are teacher-forced using ground-truth
speech timestamps. During inference, we obtain this alignment through
incremental speech synthesis. The interaction expert decodes a response
into punctuation-delimited phrases, each of which is submitted to TTS
upon completion. Speech and motion output can begin once the first
playable audio chunk is available, while subsequent phrases are decoded
and synthesized in the background.

To preserve causality, we maintain separate decoding and committed KV
caches sharing the frozen interaction-expert parameters. The decoding
cache supports response generation ahead of playback, whereas the
committed cache receives response tokens only as their corresponding
audio intervals are played. The motion expert accesses only the
committed cache, preventing future response text from entering its
interaction context. At step $t$, the committed context, robot state,
and vocal condition are used to generate the next motion chunk
$\hat{a}_{t}$, which is dispatched alongside the corresponding
speech chunk $\hat{s}_{t}$ on the shared 400-ms timeline.
Algorithm~\ref{alg:incremental_streaming} in Appendix~\ref{app:token_commitment} details
incremental generation, temporal alignment, and interruption handling.

\section{Experiments} 
\label{sec:experiments}

We conduct empirical studies to justify three central claims: (i) EIP supports natural and expressive human--robot interaction by coordinating speech with continuous, state- and emotion-aware motion; (ii) its streaming motion generation improves motion realism, speech alignment, and diversity over representative gesture-generation baselines; and (iii) compared with a cascaded variant, EIP achieves a higher observed
success rate for emotion-aware motion and lower warm-start latency,
with similar observed success rates for proactive response and
interruption-aware turn-taking.

\subsection{Experimental Setup}
\label{sec:exp_setup}

\subsubsection{Dataset Processing}
\label{sec:dataset_process}

We construct \textbf{EI-Data} by combining co-speech data from
\textbf{BEAT2} \citep{liu2024emage}, \textbf{GUAVA}
\citep{zhang2025guava}, and \textbf{SHOW} \citep{yi2023generating}
with dyadic interaction data from \textbf{Seamless Interaction}
\citep{agrawal2025seamless}. Co-speech recordings are augmented
with synthesized user turns and fixed listening segments, while
dyadic recordings preserve natural turn-taking and listening
behavior. We retarget motions to Astribot S1, filter motion
artifacts, and annotate chunk-level interaction states to provide
supervision across speaking, listening and idle. The resulting training set contains about 60,000 interaction episodes of 20 seconds each. We select 20 high-quality test episodes based on clear distinctions between speaking and listening behaviors, motion diversity, and emotion-motion consistency, while ensuring that the training and test sets are strictly disjoint. Further details on data construction, selection, and preprocessing are provided in Appendix~\ref{app:data_processing}.

\subsubsection{Implementation \& Deployment Details}
\label{sec:implementation}

The interaction expert is initialized from the pre-trained Audio-Interaction model based on Qwen2.5-3B and remains frozen throughout training. The motion expert consists of 36 Transformer blocks with 16 attention heads, a hidden dimension of \(d_m=768\), and an intermediate dimension of 3072, totaling approximately 380M parameters. We use $\Delta=0.4\mathrm{s}$ and $r_a=30\mathrm{Hz}$, giving 12 frames per motion chunk, with $N=20$ for Astribot S1 and $N=165$ for SMPL-X skeleton. Input audio is sampled at 16\,kHz, and output speech
at $r_s=24\,\mathrm{kHz}$.

\begin{figure}[!t]
    \centering
    \includegraphics[width= 0.95\linewidth]{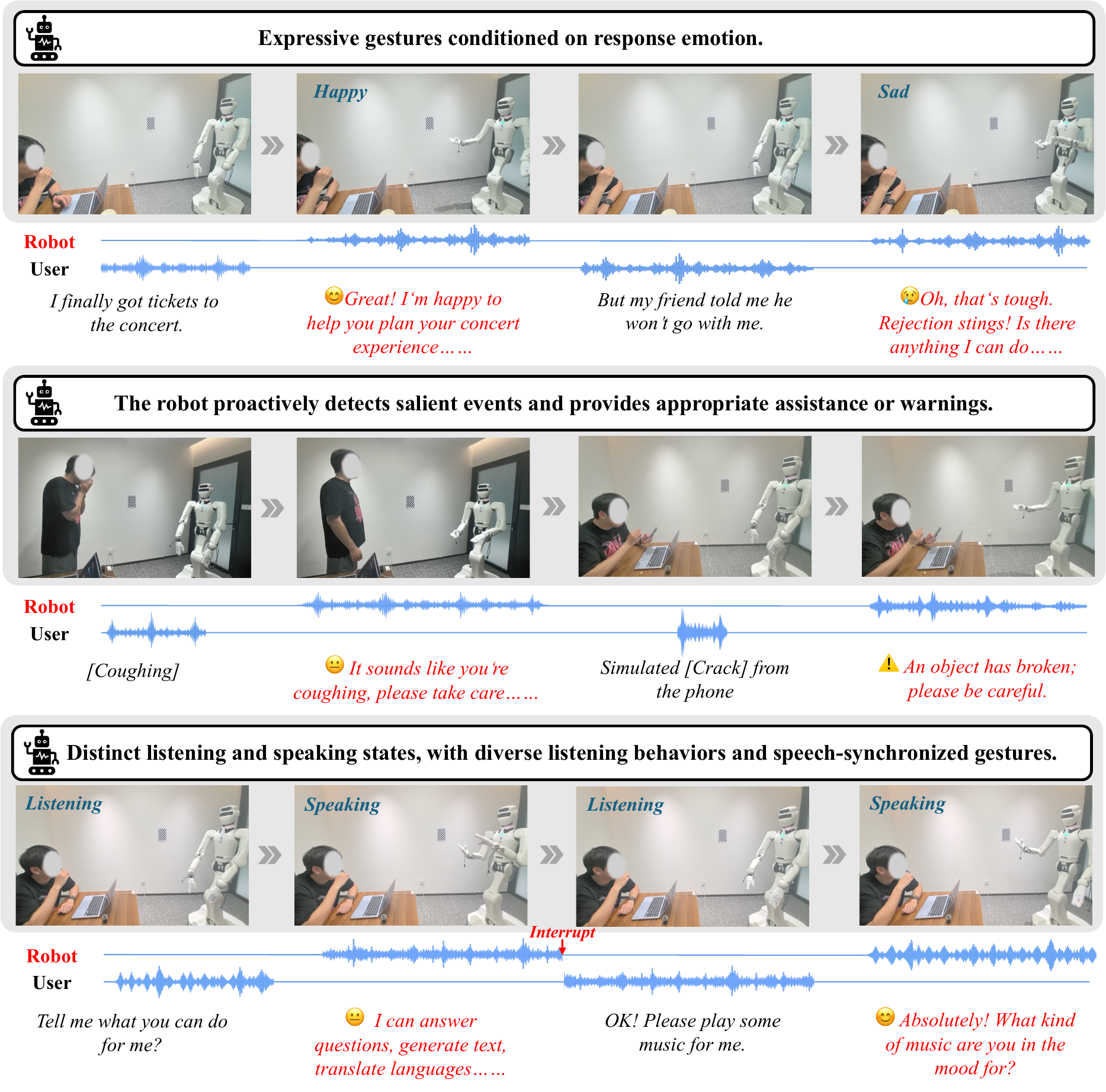}
    \caption{\textbf{Qualitative results of EIP on the physical Astribot S1.} Top: emotion-aware gesture generation. Middle: proactive responses to salient acoustic events. Bottom: interruption-aware turn-taking with distinct listening behaviors and speech-synchronized gestures.
}
    \label{fig:real result}
\end{figure}

Stage I trains the motion expert for 15K optimization steps using AdamW with a global batch size of 32. The interaction expert remains frozen, and the motion expert learns speaking, listening, and idle behaviors using only speaking-state conditioning. The loss weights are set to \(\lambda_1=1\), \(\lambda_2=2\), and \(\lambda_3=1\). A sensitivity analysis of these loss weights
is provided in Appendix~\ref{app:loss_weight_analysis}. The learning rate is warmed up for 500 steps to \(1\times10^{-4}\) and then cosine-decayed to \(1\times10^{-5}\).

Stage II starts from the Stage-I checkpoint and introduces FiLM-based emotion conditioning in the final four motion-expert blocks. The final eight motion-expert blocks and the motion head are fine-tuned for 1K steps with a global batch size of 16. Newly introduced conditioning parameters use a peak learning rate of \(1\times10^{-4}\), while the unfrozen pre-trained parameters use a \(0.02\times\) learning-rate scale. The learning rate is warmed up for 100 steps and then follows a cosine schedule. Both stages are trained on eight NVIDIA RTX A6000 GPUs.

\begin{table*}[t]
    \centering
    \caption{Observer-based user study for different training stages.
    All scores are reported on a seven-point Likert scale.}
    \label{tab:user_study_training_stages}
    \setlength{\tabcolsep}{8pt}
    \renewcommand{\arraystretch}{1}
    \begin{tabular}{lcccc|c}
        \toprule
        \multirow{2}{*}{Method}
        & \multirow{2}{*}{Naturalness $\uparrow$}
        & State
        & Listening
        & Emotional
        & \multirow{2}{*}{Average $\uparrow$} \\
        &
        & Adaptation $\uparrow$
        & Diversity $\uparrow$
        & Expressiveness $\uparrow$
        & \\
        \midrule
        a) Vanilla
        & $2.75_{\pm 2.24}$
        & $2.75_{\pm 2.08}$
        & $3.56_{\pm 2.10}$
        & $3.38_{\pm 1.50}$
        & $3.11_{\pm 0.76}$ \\
        
        b) +Stage I
        & $5.50_{\pm 1.63}$
        & $6.19_{\pm \mathbf{0.98}}$
        & $\mathbf{4.94}_{\pm \mathbf{1.69}}$
        & $4.25_{\pm 2.02}$
        & $5.22_{\pm 1.03}$ \\
        
        c) +Stage II (Final)
        & $\mathbf{6.56}_{\pm \mathbf{0.81}}$
        & $\mathbf{6.44}_{\pm 1.55}$
        & $4.88_{\pm 1.86}$
        & $\mathbf{6.06}_{\pm \mathbf{0.85}}$
        & $\mathbf{5.98}_{\pm \mathbf{0.50}}$ \\
        \bottomrule
    \end{tabular}
\end{table*}

We deploy the model on the Astribot S1, with online inference on an NVIDIA RTX 5090. The generated 30-Hz motions and synchronized speech are transmitted in 400-ms chunks to the safety gate. For collision avoidance, we apply a filter based on model predictive control (MPC) to refine the generated trajectories. A Ruckig-based online trajectory generator (OTG)~\citep{berscheid2021jerk} further enforces velocity, acceleration, and jerk limits, producing smooth commands at 50 Hz.

\subsection{Evaluation}
\label{sec:evaluation}
\subsubsection{Qualitative Results}
\label{sec:qualitative}

Fig.~\ref{fig:real result} shows EIP's behavior during real-world interactions on the Astribot S1 across three representative scenarios. EIP conditions motion generation directly on the semantic emotion of the generated response, adapting gesture expressiveness accordingly. It also proactively responds to salient acoustic events, such as coughing and the sound of an object breaking, with relevant verbal responses and accompanying gestures. During user interruptions, it transitions from speaking to listening and subsequently responds to the updated request, exhibiting distinct listening behaviors and speech-synchronized gestures. These examples illustrate coordinated verbal and non-verbal behavior throughout an interaction. Further visualizations of EIP's results in simulation are provided in Appendix~\ref{app:qualitative} (Fig.~\ref{fig:simulation_result}).

\begin{table}[t]
    \centering
    \small
    \setlength{\tabcolsep}{3pt}
    \caption{Observer-based user study results comparing RoboGesture and EIP. All scores are reported on a 7-point scale.}
    \label{tab:user-study}
    \begin{tabular*}{\columnwidth}{@{\extracolsep{\fill}}lccc@{}}
        \toprule
        Method
        & \shortstack{Motion\\Naturalness $\uparrow$}
        & \shortstack{Emotional\\Expressiveness $\uparrow$}
        & \shortstack{Gesture\\Diversity $\uparrow$} \\
        \midrule
        RoboGesture~\citep{wang2026robogesture}
        & $3.20_{\pm 1.15}$
        & $3.07_{\pm 1.10}$
        & $3.67_{\pm 1.23}$ \\
        Ours
        & $\mathbf{6.33}_{\boldsymbol{\pm 0.72}}$
        & $\mathbf{6.67}_{\boldsymbol{\pm 0.49}}$
        & $\mathbf{6.13}_{\boldsymbol{\pm 0.74}}$ \\
        \bottomrule
    \end{tabular*}
    \vspace{-4pt}
\end{table}

\subsubsection{Quantitative User Study}
\label{sec:user_study}

\begin{wrapfigure}{r}{0.5\textwidth}
    \centering
    \vspace{-10pt}
    \includegraphics[width=\linewidth]{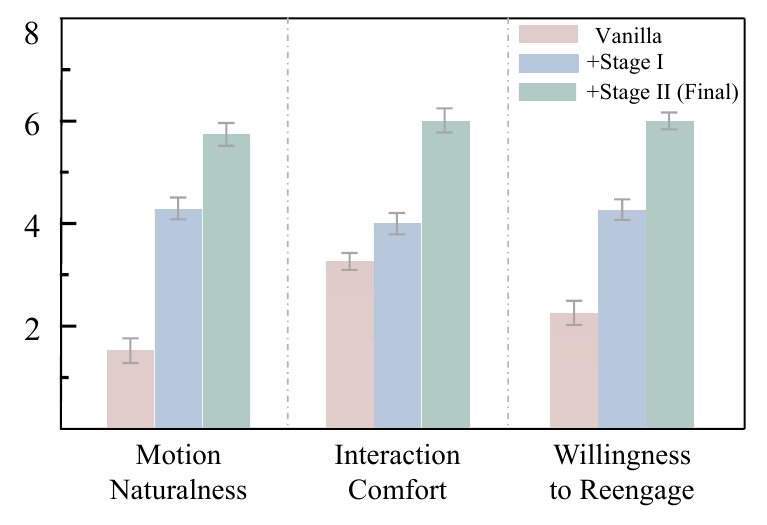}
    \caption{
        Participant-based user study on the physical robot.
        Twelve participants rated Vanilla, +Stage I, and +Stage II (Final)
      on a seven-point Likert scale using Motion Naturalness,
        Interaction Comfort, and Willingness to Re-engage.
        Bars show mean ratings, and error bars indicate standard deviations.
    }
    \label{fig:real user study}
    \vspace{-8pt}
\end{wrapfigure}

We evaluate EIP from two complementary perspectives: observer-based evaluation of offline-generated videos and participant-based evaluation with the physical robot. In the observer-based study, 16 participants rated three configurations on a seven-point Likert scale: \textbf{Vanilla}, which uses the same motion backbone and training data, is conditioned only on the interaction context $h_t$ and trained with the standard flow-matching objective, without vocal-token conditioning, phase-specific supervision; \textbf{+Stage I}, which adds phase-specific supervision for speaking, listening, and idle states; and \textbf{+Stage II (Final)}, which further introduces FiLM-based emotion conditioning. We evaluate Motion Naturalness, State Adaptation, Listening Diversity, and Emotional Expressiveness, measuring motion realism, state-appropriate behavior, listening-gesture variety, and emotion-motion alignment, respectively. In the physical-robot study, 12 participants rated Motion Naturalness, Interaction Comfort, and Willingness to Re-engage.

As shown in Table~\ref{tab:user_study_training_stages}, Stage I substantially improves the average score from 3.11 to 5.22, particularly for Motion Naturalness and State Adaptation. Stage II further increases the average to 5.98 and improves Emotional Expressiveness from 4.25 to 6.06, while largely preserving the interaction behaviors learned in Stage I. Figure~\ref{fig:real user study} shows consistent improvements in the physical setting, where the final model achieves the highest scores on all three metrics. These results indicate that the two-stage training strategy improves both motion quality and the overall comfort and engagement of human--robot interaction.

Table~\ref{tab:user-study} shows that EIP receives higher mean ratings
than RoboGesture for Motion Naturalness, Emotional Expressiveness,
and Gesture Diversity. The comparison settings are detailed in
Appendix~\ref{app:qualitative}.

\subsubsection{Quantitative Streaming Motion Evaluation}
\label{sec:motion_eval}

We compare EIP with MIBURI~\citep{mughal2026miburi},
EMAGE~\citep{liu2024emage}, Photoreal~\citep{ng2024photoreal},
and RoboGesture~\citep{wang2026robogesture} on the EI-Data
test set using Fr\'echet Gesture Distance (FGD), Beat Consistency
(BC), and Diversity (Div). These baselines generate motion
conditioned on reference target-speaker speech. To isolate
motion generation from response-generation errors, we condition
EIP on user audio and teacher-forced target text for each
predicted motion chunk. As shown in
Table~\ref{tab:quantitative_results}, EIP achieves the best results among all evaluated methods
across all three metrics on the Astribot S1, reducing FGD
from 74.51 to 43.83 compared with the strongest baseline.
Additional results on the SMPL-X skeleton are provided in
Appendix~\ref{app:qualitative}, where EIP also
outperforms all evaluated baselines. Further ablations validate our architectural and training choices. \textbf{Ours (w/o $\mathbf{h}_{t}$)} blocks
$\mathbf{Q}_t, \mathbf{v}_t$ and $\mathbf{A}^{\tau}_t$ from attending to $\mathbf{h}_{t}$,
its poorer performance relative to \textbf{Ours (Final)} supports the effectiveness
of interaction-expert hidden states as context for robot motion
generation. \textbf{Final-layer Broadcast} (FGD 61.21), which replaces layer-wise MoT coupling with broadcasting the final interaction-layer features to all motion layers, and \textbf{Cascade} (65.39), which conditions the motion expert on response speech features extracted by an independent audio encoder via cross-attention, both underperform our layer-wise joint attention, highlighting the benefit of multi-level interaction--motion coupling. In both Tables ~\ref{tab:quantitative_results} and ~\ref{tab:interaction_results}, Cascade shares EIP’s training data, motion backbone, vocal/FiLM conditioning, two-stage training, and incremental TTS scheduling. \textbf{Bi-Chunk Attn.}, which allows mutual attention among $Q_t$, $v_t$, and $A_t$ within each chunk, increases FGD to 52.17, suggesting that preserving the dependency structure between conditioning tokens and noisy motion tokens improves generation quality. Finally, \textbf{Joint Training} increases FGD to 90.73, supporting our two-stage strategy of learning state adaptation before introducing emotion conditioning, thereby reducing interference between the two learning objectives.

\subsubsection{Interaction Success Rate and Latency}
\label{sec:interaction_eval}

\begin{wraptable}{r}{0.48\textwidth}
    \centering
    \vspace{-12pt}
    \caption{
    Quantitative motion evaluation on the Astribot S1.
    For baseline methods, $^\ast$ denotes training on our
    EI-Data in the target skeleton; otherwise, generated
    motions are directly retargeted to the target skeleton
    for evaluation.
    }
    \label{tab:quantitative_results}
    \small
    \setlength{\tabcolsep}{5pt}
    \begin{tabular}{@{}lccc@{}}
        \toprule
        System
        & FGD$\downarrow$
        & BC$\to$
        & Div$\uparrow$ \\
        \midrule
        GT
        & -- & 0.780 & 1.118 \\
        \midrule
        MIBURI~\citep{mughal2026miburi}
        & 137.74 & 0.666 & 0.910 \\

        EMAGE~\citep{liu2024emage}
        & 174.39 & 0.689 & 0.912 \\

        Photoreal~\citep{ng2024photoreal}
        & 138.03 & 0.665 & 0.927 \\

        RoboGesture~\citep{wang2026robogesture}
        & 198.71 & 0.513 & 0.936 \\

        MIBURI$^\ast$~\citep{mughal2026miburi}
        & 74.51 & 0.697 & 1.007 \\

        RoboGesture$^\ast$~\citep{wang2026robogesture}
        & 92.99 & 0.654 & 0.967 \\
        \midrule
        Vanilla
        & 126.84 & 0.601 & 0.948 \\

        Ours (w/o $\mathbf{h}_{t}$)
        & 90.27 & 0.677 & 0.931 \\

        Ours (Joint Training)
        & 90.73 & 0.710 & 0.921 \\

        Ours (Cascade)
        & 65.39 & 0.727 & 0.994 \\

        Ours (Final-layer Broadcast)
        & 61.21 & 0.711 & 0.980 \\

        Ours (Bi-Chunk Attn.)
        & 52.17 & 0.740 & 0.982 \\

        \textbf{Ours (Final)}
        & \textbf{43.83}
        & \textbf{0.744}
        & \textbf{1.021} \\
        \bottomrule
    \end{tabular}
    \vspace{-6pt}
\end{wraptable}

We compare EIP with RoboGesture and our cascaded variant across three interaction scenarios: proactive response, interruption-aware turn-taking, and emotion-aware motion, using 20 AI-generated speech samples per scenario. Success is assessed by human evaluators based on proactive event responsiveness (e.g., coughing, alarms, falls, or thunder), interruption handling, and emotion-consistent motion, respectively.
In the interruption scenario, interruptions occur at randomly
sampled delays of 1--5 seconds after the end of the preceding
speech segment. Warm-start latency measures the time from speech input to the first motion chunk's execution onset. RTF is the ratio of end-to-end processing time to generated motion duration. As shown in Table~\ref{tab:interaction_results}, EIP
achieves an average success rate of 91.7\%, outperforming
Cascade (76.7\%) and RoboGesture (18.3\%). While Cascade performs comparably in proactive response (90\%) and interruption-aware turn-taking (85\% vs. 90\%), EIP achieves higher emotion-aware motion success (95\% vs. 55\%), highlighting the benefit of direct interaction-motion coupling. EIP also reduces latency from 1.99 s to 1.35 s
and RTF from 0.952 to 0.804 compared with Cascade, while
achieving lower latency than RoboGesture (3.78 s).

\begin{table*}[!h]
    \centering
    \caption{\textbf{Interaction success rate and inference efficiency.}
EIP achieves the highest average success rate and the lowest
latency and RTF across the evaluated methods.
All evaluations are conducted on an NVIDIA RTX A6000 GPU.}
    \label{tab:interaction_results}
    \renewcommand{\arraystretch}{1}
    \setlength{\tabcolsep}{4pt}
    \small
    \begin{tabular}{lccc|ccc}
    \toprule
    \textbf{Method} &
    \makecell{\textbf{Proactive Event}\\\textbf{Response (\%) $\uparrow$}} &
    \makecell{\textbf{Interruption-Aware}\\\textbf{Turn Taking (\%) $\uparrow$}} &
    \makecell{\textbf{Emotion-Aware}\\\textbf{Motion (\%) $\uparrow$}} &
    \makecell{\textbf{Average}\\\textbf{Success (\%) $\uparrow$}} &
    \makecell{\textbf{Latency (s)}\\$\downarrow$} &
    \makecell{\textbf{RTF}\\$\downarrow$} \\
    \midrule
    RoboGesture~\citep{wang2026robogesture} & 0 & 15 & 40 & 18.3 & 3.78 & 0.821 \\
    Ours (Cascade) & \textbf{90} & 85 & 55 & 76.7 & 1.99 & 0.952 \\
    Ours &
    \textbf{90} & \textbf{90} & \textbf{95} &
    \textbf{91.7} & \textbf{1.35} & \textbf{0.804} \\
    \bottomrule
    \end{tabular}
\end{table*}

\section{Conclusion} 
\label{sec:conclusion}

We present Embodied Interaction Policy (EIP), a unified streaming framework coupling interaction modeling and motion generation on a shared 400-ms timeline. Building on a frozen pretrained interaction model, EIP learns continuous
robot motion that adapts to conversational state and response emotion
through two-stage motion training. We also introduced EI-Data to support policy learning and evaluate
motion quality. Quantitative evaluations, user studies, and deployment on the Astribot S1 demonstrate improved motion quality, higher interaction
success rates, reduced response latency, and greater interaction comfort and engagement. Despite these promising results, EIP relies on audio to perceive interaction partners, limiting its ability to interpret visual cues such as gaze, facial expressions, and body posture. Future work will incorporate visual perception to capture these non-verbal cues and support more context-aware embodied interaction.

% \input{samplebody-journals}

% Bibliography
\bibliographystyle{ACM-Reference-Format}
\bibliography{main}

@String(CVPR= {IEEE Conf. Comput. Vis. Pattern Recog.})

@String(ICCV= {Int. Conf. Comput. Vis.})

@String(AAAI = {AAAI})

@String(CVPR  = {CVPR})

@String(ICCV  = {ICCV})

@article{xie2026audio,
  title={Audio Interaction Model},
  author={Xie, Zhifei and Liu, Zihang and An, Ze and Hu, Xiaobin and Liao, Yue and Ma, Ziyang and Yang, Dongchao and Lin, Mingbao and Ye, Deheng and Yan, Shuicheng and others},
  journal={arXiv preprint arXiv:2606.05121},
  year={2026}
}

@inproceedings{zhou2025transfusion,
  title={Transfusion: Predict the next token and diffuse images with one multi-modal model},
  author={Zhou, Chunting and Yu, Lili and Babu, Arun and Tirumala, Kushal and Yasunaga, Michihiro and Shamis, Leonid and Kahn, Jacob and Ma, Xuezhe and Zettlemoyer, Luke and Levy, Omer},
  booktitle={International Conference on Learning Representations},
  volume={2025},
  pages={6446--6469},
  year={2025}
}

@article{black2024pi_0,
  title = {{$\pi_0$}: A Vision-Language-Action Flow Model for General Robot Control},
  author={Black, Kevin and Brown, Noah and Driess, Danny and Esmail, Adnan and Equi, Michael and Finn, Chelsea and Fusai, Niccolo and Groom, Lachy and Hausman, Karol and Ichter, Brian and others},
  journal={arXiv preprint arXiv:2410.24164},
  year={2024}
}

@inproceedings{liu2024emage,
  title={Emage: Towards unified holistic co-speech gesture generation via expressive masked audio gesture modeling},
  author={Liu, Haiyang and Zhu, Zihao and Becherini, Giorgio and Peng, Yichen and Su, Mingyang and Zhou, You and Zhe, Xuefei and Iwamoto, Naoya and Zheng, Bo and Black, Michael J},
  booktitle={2024 IEEE/CVF Conference on Computer Vision and Pattern Recognition (CVPR)},
  pages={1144--1154},
  year={2024},
  organization={IEEE}
}

@inproceedings{zhang2025guava,
  title={Guava: Generalizable upper body 3d gaussian avatar},
  author={Zhang, Dongbin and Liu, Yunfei and Lin, Lijian and Zhu, Ye and Li, Yang and Qin, Minghan and Li, Yu and Wang, Haoqian},
  booktitle={2025 IEEE/CVF International Conference on Computer Vision (ICCV)},
  pages={14205--14217},
  year={2025},
  organization={IEEE}
}

@inproceedings{yi2023generating,
  title={Generating holistic 3d human motion from speech},
  author={Yi, Hongwei and Liang, Hualin and Liu, Yifei and Cao, Qiong and Wen, Yandong and Bolkart, Timo and Tao, Dacheng and Black, Michael J},
  booktitle={2023 IEEE/CVF Conference on Computer Vision and Pattern Recognition (CVPR)},
  pages={469--480},
  year={2023},
  organization={IEEE}
}

@article{agrawal2025seamless,
  title={Seamless interaction: Dyadic audiovisual motion modeling and large-scale dataset},
  author={Agrawal, Vasu and Akinyemi, Akinniyi and Alvero, Kathryn and Behrooz, Morteza and Buffalini, Julia and Carlucci, Fabio Maria and Chen, Joy and Chen, Junming and Chen, Zhang and Cheng, Shiyang and others},
  journal={arXiv preprint arXiv:2506.22554},
  year={2025}
}

@inproceedings{pavlakos2019expressive,
  title={Expressive body capture: 3d hands, face, and body from a single image},
  author={Pavlakos, Georgios and Choutas, Vasileios and Ghorbani, Nima and Bolkart, Timo and Osman, Ahmed AA and Tzionas, Dimitrios and Black, Michael J},
  booktitle={Proceedings of the IEEE/CVF conference on computer vision and pattern recognition},
  pages={10975--10985},
  year={2019}
}

@article{saito2026soma,
  title={Soma: Unifying parametric human body models},
  author={Saito, Jun and Li, Jiefeng and de Ruyter, Michael and Guerrero, Miguel and Lim, Edy and Hassani, Ehsan and Ribera, Roger Blanco and Moon, Hyejin and Dadela, Magdalena and Di Lucca, Marco and others},
  journal={arXiv preprint arXiv:2603.16858},
  year={2026}
}

@misc{warp2022,
  title        = {Warp: A High-performance Python Framework for GPU Simulation and Graphics},
  author       = {Macklin, Miles},
  month        = mar,
  year         = {2022},
  note         = {NVIDIA GPU Technology Conference (GTC)},
  howpublished = {\url{https://github.com/NVIDIA/warp}}
}

@misc{qwen36_27b,
    title = {{Qwen3.6-27B}: Flagship-Level Coding in a 27B Dense Model},
    url = {https://qwen.ai/blog?id=qwen3.6-27b},
    author = {{Qwen Team}},
    month = {April},
    year = {2026}
}

@article{hu2026qwen3,
  title={Qwen3-tts technical report},
  author={Hu, Hangrui and Zhu, Xinfa and He, Ting and Guo, Dake and Zhang, Bin and Wang, Xiong and Guo, Zhifang and Jiang, Ziyue and Hao, Hongkun and Guo, Zishan and others},
  journal={arXiv preprint arXiv:2601.15621},
  year={2026}
}

@inproceedings{yao2023react,
  title     = {ReAct: Synergizing Reasoning and Acting in Language Models},
  author    = {Yao, Shunyu and Zhao, Jeffrey and Yu, Dian and Du, Nan
               and Shafran, Izhak and Narasimhan, Karthik R. and Cao, Yuan},
  booktitle = {International Conference on Learning Representations},
  year      = {2023}
}

@inproceedings{schick2023toolformer,
  title     = {Toolformer: Language Models Can Teach Themselves to Use Tools},
  author    = {Schick, Timo and Dwivedi-Yu, Jane and Dess{\`i}, Roberto
               and Raileanu, Roberta and Lomeli, Maria and Hambro, Eric
               and Zettlemoyer, Luke and Cancedda, Nicola and Scialom, Thomas},
  booktitle = {Advances in Neural Information Processing Systems},
  volume    = {36},
  pages     = {68539--68551},
  year      = {2023}
}

@article{defossez2024moshi,
  title   = {Moshi: A Speech-Text Foundation Model for Real-Time Dialogue},
  author  = {D{\'e}fossez, Alexandre and Mazar{\'e}, Laurent and Orsini, Manu
             and Royer, Am{\'e}lie and P{\'e}rez, Patrick and J{\'e}gou, Herv{\'e}
             and Grave, Edouard and Zeghidour, Neil},
  journal = {arXiv preprint arXiv:2410.00037},
  year    = {2024}
}

@inproceedings{ma2025language,
  title     = {Language Model Can Listen While Speaking},
  author    = {Ma, Ziyang and Song, Yakun and Du, Chenpeng and Cong, Jian
               and Chen, Zhuo and Wang, Yuping and Wang, Yuxuan and Chen, Xie},
  booktitle = {Proceedings of the AAAI Conference on Artificial Intelligence},
  volume    = {39},
  number    = {23},
  pages     = {24831--24839},
  year      = {2025},
  doi       = {10.1609/aaai.v39i23.34665}
}

@article{xu2025qwen25omni,
  title   = {Qwen2.5-Omni Technical Report},
  author  = {Xu, Jin and Guo, Zhifang and He, Jinzheng and Hu, Hangrui
             and He, Ting and Bai, Shuai and Chen, Keqin and Wang, Jialin
             and Fan, Yang and Dang, Kai and Zhang, Bin and Wang, Xiong
             and Chu, Yunfei and Lin, Junyang},
  journal = {arXiv preprint arXiv:2503.20215},
  year    = {2025}
}

@article{thinkingmachines2026interactionmodels,
  author  = {{Thinking Machines Lab}},
  title   = {Interaction Models: A Scalable Approach to Human-AI Collaboration},
  journal = {Thinking Machines Lab: Connectionism},
  year    = {2026},
  month   = {May},
  doi     = {10.64434/tml.20260511}
}

@inproceedings{ng2024photoreal,
  title     = {From Audio to Photoreal Embodiment:
               Synthesizing Humans in Conversations},
  author    = {Ng, Evonne and Romero, Javier and Bagautdinov, Timur
               and Bai, Shaojie and Darrell, Trevor and Kanazawa, Angjoo
               and Richard, Alexander},
  booktitle = {Proceedings of the IEEE/CVF Conference on Computer Vision
               and Pattern Recognition},
  pages     = {1001--1010},
  year      = {2024}
}

@misc{wang2026robogesture,
  title         = {{RoboGesture}: Real-Time Semantic-aligned Co-Speech
                   Gestures Generation for Humanoid Interaction},
  author        = {Wang, Zifan and Ren, Ziang and Shi, Pengyang
                   and Wang, Zirui and Lin, Chenghuai and Wang, Tianze
                   and Qi, Zekun and Zhao, Liangliang and Wang, He
                   and Yi, Li},
  year          = {2026},
  eprint        = {2608.28693},
  archivePrefix = {arXiv},
  primaryClass  = {cs.RO}
}

@article{zhang2026proact,
  title   = {{ProAct}: Harnessing Streaming Motion Generation and Agentic
             Reasoning for Real-Time Embodied Social Interaction},
  author  = {Zhang, Zeyi and Kang, Zixi and Zhao, Ruijie and Feng, Yusen
             and Jiang, Biao and Ji, Hanyu and Liu, Libin},
  journal = {ACM Transactions on Graphics},
  volume  = {45},
  number  = {6},
  year    = {2026},
  doi     = {10.1145/3842576}
}

@inproceedings{cai2024digital,
  title     = {Digital Life Project: Autonomous 3D Characters with Social Intelligence},
  author    = {Cai, Zhongang and Jiang, Jianping and Qing, Zhongfei and Guo, Xinying
               and Zhang, Mingyuan and Lin, Zhengyu and Mei, Haiyi and Wei, Chen
               and Wang, Ruisi and Yin, Wanqi and Pan, Liang and Fan, Xiangyu
               and Du, Han and Gao, Peng and Yang, Zhitao and Gao, Yang
               and Li, Jiaqi and Ren, Tianxiang and Wei, Yukun and Wang, Xiaogang
               and Loy, Chen Change and Yang, Lei and Liu, Ziwei},
  booktitle = {Proceedings of the IEEE/CVF Conference on Computer Vision and Pattern Recognition},
  pages     = {582--592},
  year      = {2024}
}

@inproceedings{jiang2025solami,
  title     = {{SOLAMI}: Social Vision-Language-Action Modeling for Immersive
               Interaction with 3D Autonomous Characters},
  author    = {Jiang, Jianping and Xiao, Weiye and Lin, Zhengyu and Zhang, Huaizhong
               and Ren, Tianxiang and Gao, Yang and Lin, Zhiqian and Cai, Zhongang
               and Yang, Lei and Liu, Ziwei},
  booktitle = {Proceedings of the IEEE/CVF Conference on Computer Vision and Pattern Recognition},
  pages     = {26887--26898},
  year      = {2025}
}

@inproceedings{yoon2019robots,
  title     = {Robots Learn Social Skills: End-to-End Learning of Co-Speech
               Gesture Generation for Humanoid Robots},
  author    = {Yoon, Youngwoo and Ko, Woo-Ri and Jang, Minsu and Lee, Jaeyeon
               and Kim, Jaehong and Lee, Geehyuk},
  booktitle = {2019 International Conference on Robotics and Automation (ICRA)},
  pages     = {4303--4309},
  year      = {2019},
  doi       = {10.1109/ICRA.2019.8793720}
}

@inproceedings{araujo2026retargeting,
  title     = {Retargeting Matters: General Motion Retargeting for Humanoid
               Motion Tracking},
  author    = {Araujo, Joao Pedro and Ze, Yanjie and Xu, Pei and Wu, Jiajun
               and Liu, C. Karen},
  booktitle = {2026 IEEE International Conference on Robotics and Automation (ICRA)},
  year      = {2026}
}

@inproceedings{yang2026omniretarget,
  title     = {{OmniRetarget}: Interaction-Preserving Data Generation for
               Humanoid Whole-Body Loco-Manipulation and Scene Interaction},
  author    = {Yang, Lujie and Huang, Xiaoyu and Wu, Zhen and Kanazawa, Angjoo
               and Abbeel, Pieter and Sferrazza, Carmelo and Liu, C. Karen
               and Duan, Rocky and Shi, Guanya},
  booktitle = {2026 IEEE International Conference on Robotics and Automation (ICRA)},
  year      = {2026}
}

@article{mughal2026miburi,
  title={Miburi: Towards expressive interactive gesture synthesis},
  author={Mughal, M Hamza and Dabral, Rishabh and Demberg, Vera and Theobalt, Christian},
  journal={arXiv preprint arXiv:2603.03282},
  year={2026}
}

@article{skantze2021turntaking,
  title={Turn-taking in Conversational Systems and Human-Robot Interaction: A Review},
  author={Skantze, Gabriel},
  journal={Computer Speech \& Language},
  volume={67},
  pages={101178},
  year={2021},
  doi={10.1016/j.csl.2020.101178}
}

@inproceedings{berscheid2021jerk,
  title={Jerk-limited Real-time Trajectory Generation with Arbitrary Target States},
  author={Berscheid, Lars and Kr{\"o}ger, Torsten},
  booktitle={Robotics: Science and Systems},
  year={2021},
  doi={10.15607/RSS.2021.XVII.015}
}

\clearpage 
\appendix
\phantomsection
\begin{center}
{\LARGE\sc Supplementary Materials}
\end{center}

This appendix provides additional material that complements the main paper.
Section~\ref{app:data_processing} describes the construction of EI-Data, covering motion normalization and robot retargeting (Section~\ref{app:motion_retargeting}), real-interaction episodes from Seamless Interaction (Section~\ref{app:seamless_processing}), synthetic-context co-speech episodes (Section~\ref{app:cospeech_processing}), and emotion annotation for second-stage training (Section~\ref{app:emotion_annotation}).
Section~\ref{app:token_commitment} specifies playback-synchronized token commitment (Algorithm~\ref{alg:incremental_streaming}), which aligns response-token KV-cache updates with speech playback. Section~\ref{app:mask_rationale} explains the rationale behind the token ordering and attention visibility in the causal mask. Section~\ref{app:loss_weight_analysis} provides a sensitivity analysis of the Stage-I loss weights.
Section~\ref{sec:emotion_ablation} ablates the contributions of emotion embeddings and FiLM conditioning.
Section~\ref{app:qualitative} presents additional qualitative results of EIP in simulation (Fig.~\ref{fig:simulation_result}).

\section{Data Processing Details}
\label{app:data_processing}

We construct two types of training episodes: real interactions from Seamless Interaction \citep{agrawal2025seamless}, and synthetic-context co-speech episodes derived from BEAT2 \citep{liu2024emage}, GUAVA \citep{zhang2025guava}, and SHOW \citep{yi2023generating}.
Throughout the data-processing pipeline, we refer to the person whose motion is retargeted to the robot as the \emph{target}, and to the other participant whose speech is provided as input as the \emph{user} or \emph{partner}.

\subsection{Motion Normalization and Robot Retargeting}
\label{app:motion_retargeting}

\begin{figure*}[h]
    \centering
    \includegraphics[width=\textwidth]{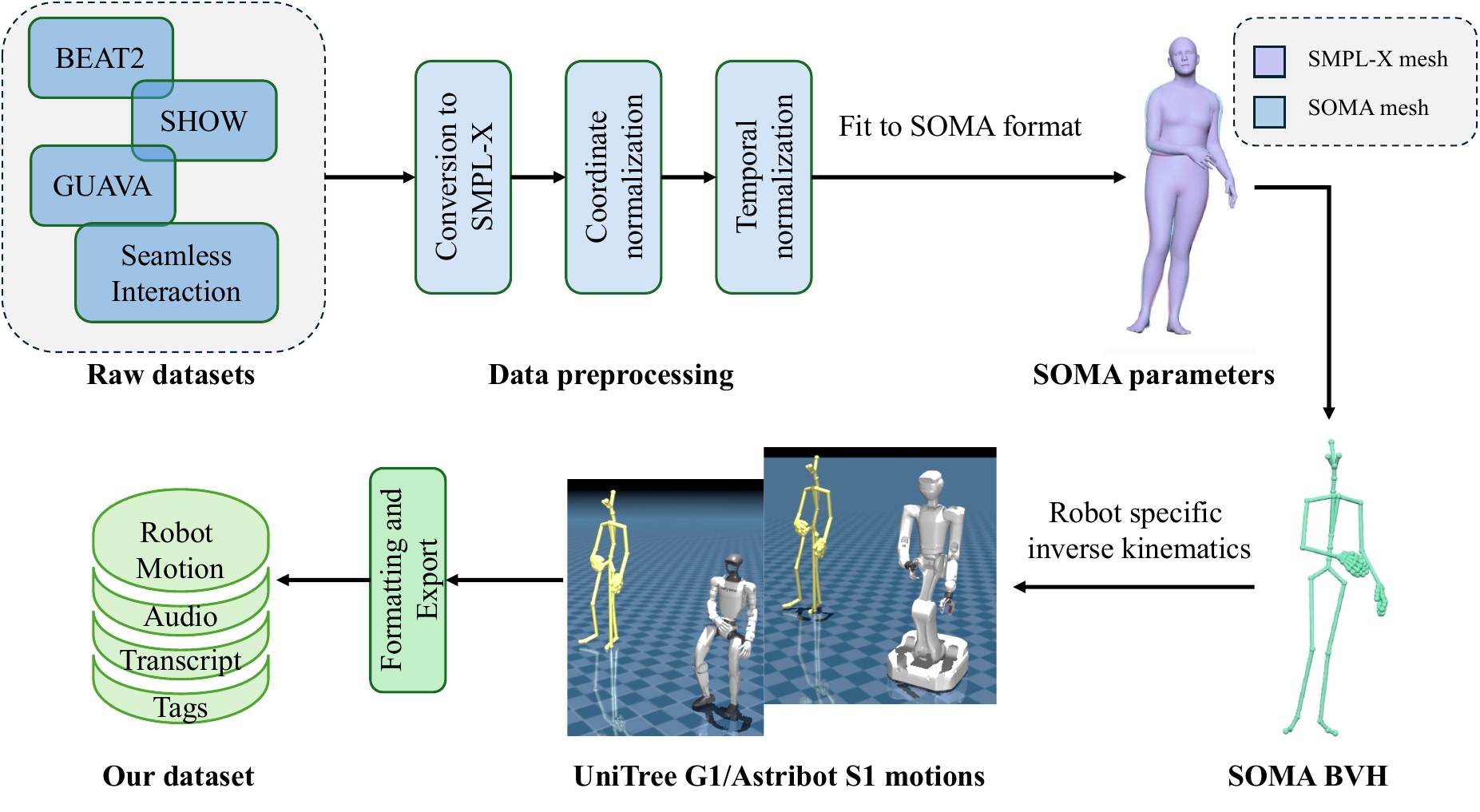}
    \caption{
        \textbf{Motion normalization and robot-retargeting pipeline.}
        Human motion is converted to a common SMPL-X representation,
        temporally normalized, fitted to SOMA, and retargeted to
        the robot via inverse kinematics.
    }
    \label{fig:retargeting}
\end{figure*}

As illustrated in Figure~\ref{fig:retargeting}, we process the human motion in the following stages.

\paragraph{Conversion to a common SMPL-X representation.}
Dataset-specific loaders first convert all motion sequences into a common SMPL-X representation \citep{pavlakos2019expressive} consisting of the global root orientation, body pose, root translation, and body-shape parameters.
BEAT2 \citep{liu2024emage} stores each motion frame using a dataset-specific 215-dimensional parameterization containing facial expression, jaw and eye poses, global orientation, body pose, and bilateral hand poses.
GUAVA \citep{zhang2025guava}, SHOW \citep{yi2023generating}, and Seamless Interaction \citep{agrawal2025seamless} are loaded from AMASS-style SMPL-X files.
We preserve the available hand, facial, and expression parameters, although the robot supervision primarily uses body and arm motion.

\paragraph{Coordinate and temporal normalization.}
We express the human motion in meters using a canonical Y-up coordinate system.
For sequences with a dataset-specific global orientation offset, we remove the sequence-level mean root rotation and select the axis-aligned world rotation that makes the body upright.
The same world rotation is applied to both the global orientation and root translation.
For GUAVA, we apply a five-frame sliding rotation average on $\mathrm{SO}(3)$ to suppress high-frequency pose jitter.

To reduce the cost of fitting, the input sequence is temporally subsampled by a factor of two while its frame-rate metadata is adjusted accordingly, preserving its physical duration.
We then resample all normalized sequences to $50$~Hz.
Rotations are interpolated with spherical linear interpolation, while translations and other Euclidean quantities are interpolated linearly.

\paragraph{SMPL-X-to-SOMA fitting.}
We first use the SMPL-X parameters to reconstruct the posed human mesh.
SOMA-X is then used to fit the intermediate SOMA skeleton to the mesh vertices \citep{saito2026soma}.
Shape-dependent identity parameters are prepared once for each body shape and reused across frames.
Fitting is performed in batches of 128 frames using two body-fitting iterations, one full-body iteration, and three Lie-algebra refinement iterations.
The resulting intermediate representation contains the SOMA joint rotations, root translation, and per-vertex fitting error.
We also export the fitted motion as a SOMA BVH sequence, which serves as the common input to the robot-specific retargeters.

\paragraph{Robot-specific inverse kinematics.}
The SOMA motion is retargeted independently to the Unitree G1 and Astribot S1 embodiments.
We first rescale the human skeleton using robot-specific segment-length factors and transform the motion into the robot coordinate system.
A Newton-based inverse-kinematics solver, accelerated with NVIDIA Warp \citep{warp2022}, then optimizes the robot configuration at every frame.

For G1, the IK objectives constrain the pelvis, torso, shoulders, elbows, wrists, knees, and feet while respecting joint limits and encouraging temporal smoothness.
Foot stabilization is applied as a post-processing step.
For S1, the planar chassis pose is obtained from the projected SOMA hip translation and heading, while the upper-body IK objectives match the shoulder, elbow, and wrist trajectories.
Head yaw and pitch are recovered from the relative chest--neck orientation and constrained to the physical head-joint range.
We use 24 IK iterations for G1 and 40 iterations for S1.
Long sequences are processed in contiguous temporal blocks while carrying the final joint state of one block into the next, preventing discontinuities at block boundaries.

\paragraph{Motion representation used for training.}
The retargeter exports robot joint trajectories at $50$~Hz.
For S1, the stored trajectory contains the planar chassis, torso, head, two arms, and grippers.
Before training, we convert angular quantities from degrees to radians, resample the trajectory to $30$~Hz, and retain 20 actuated upper-body degrees of freedom: four torso joints, two head joints, and seven joints for each arm.
The planar chassis and gripper joints are not included in the predicted motion vector.

Each $0.4$-s interval consequently contains 12 robot frames.
At streaming step $t$, the motion in the current interval is used as the robot state, and the following 12-frame interval is used as the target motion:
\begin{equation}
    q_t = q^{\mathrm{seq}}_{12t:12(t+1)}, \quad
    a_t = q^{\mathrm{seq}}_{12(t+1):12(t+2)},
    \label{eq:state_action_alignment}
\end{equation}
where $q^{\mathrm{seq}} \in \mathbb{R}^{T \times 20}$
denotes the S1 joint trajectory.

\subsection{Real-Interaction Episodes from Seamless Interaction}
\label{app:seamless_processing}

Seamless Interaction \citep{agrawal2025seamless} provides recordings of two interacting participants.
We retain the original speech of the partner and the retargeted motion of the target, thereby preserving real turn-taking, interruptions, pauses, and listening behavior.
The processing is performed as follows.

\paragraph{Participant pairing.}
We group recordings by their shared interaction identifier and match each retargeted target sequence with the unique other participant in the same interaction.

A valid pair must contain the audio, transcript, and video streams of both participants, together with the target robot motion.
Because the construction is directional, either participant can serve as the target when a valid retargeted motion sequence is available.

\paragraph{Common valid interval.}
We compute the usable duration as the minimum duration among the two audio streams, the two videos, and the target motion.
The sequence is further truncated at the last detected speech event so that long, inactive tails do not dominate training.

\paragraph{Chunk-level speech states.}
Voice-activity intervals are read from the dataset annotations, and neighboring intervals separated by at most $1$~s are merged into a single utterance.
The synchronized streams are divided into non-overlapping $0.4$-s chunks.
For each participant, every chunk is assigned one of three states:
\begin{equation}
    z_k \in
    \{\texttt{onset},\texttt{speaking},\texttt{non\_speech}\}.
\end{equation}
A chunk is labeled \texttt{onset} if it contains the beginning of an utterance, \texttt{speaking} if it overlaps an utterance without its onset, and \texttt{non\_speech} otherwise.
Response-onset supervision applies to \texttt{onset} and \texttt{non\_speech} chunks; chunks in which the target is already speaking are excluded from this loss because the response has already been initiated.
Word-level transcript tokens are assigned to chunks according to their timestamps.

\paragraph{Partner-first episode extraction.}
Candidate episodes are anchored at a partner-speech onset.
Each episode begins at the anchor and spans $20$~s:
\begin{equation}
\begin{aligned}
    T_{\mathrm{episode}} &= 20~\mathrm{s}, \qquad
    \Delta t = 0.4~\mathrm{s}, \\
    N_{\mathrm{chunk}} &=
    \frac{T_{\mathrm{episode}}}{\Delta t} = 50.
\end{aligned}
\end{equation}
We retain only complete 50-chunk windows containing a subsequent target response and at least one valid target transcript token.
Accepted windows from the same interaction are made non-overlapping.
This provides the model with both the user's speech turn and the target's transition from listening to speaking.

\paragraph{Quality filtering and final subset.}
We discard episodes with missing modalities, incomplete chunk sequences, invalid transcripts, or unavailable robot motion.
We additionally measure joint velocity, acceleration, and within-chunk motion range during target-non-speech intervals.
Sustained-motion statistics are constrained by a one-sided 90th-percentile envelope, and spike-related statistics by a 95th-percentile envelope, estimated from a manually inspected clean subset.
This filtering removes retargeting failures and implausible motion spikes while retaining naturally moving listeners.
We deterministically select 30,000 complete episodes, corresponding to 1,500,000 synchronized $0.4$-s chunks.

\subsection{Synthetic-Context Co-Speech Episodes}
\label{app:cospeech_processing}

\begin{wraptable}{r}{0.4\textwidth}
    \centering
    \caption{Number of synthetic-context co-speech episodes.}
    \setlength{\tabcolsep}{10pt}
    \label{tab:cospeech_stats}
    \begin{tabular}{lr}
        \toprule
        Source dataset & Number of episodes \\
        \midrule
        BEAT2 & 9,242 \\
        GUAVA & 3,858 \\
        SHOW  & 15,937 \\
        \midrule
        Total & 29,037 \\
        \bottomrule
    \end{tabular}
\end{wraptable}

BEAT2 \citep{liu2024emage}, GUAVA \citep{zhang2025guava}, and SHOW \citep{yi2023generating} contain a target speaker's speech, transcript, and co-speech motion, but do not provide the preceding user turn.
We therefore construct a synthetic interaction context for each usable target clip.

\paragraph{Preceding-user-utterance generation.}
Given the complete target transcript, Qwen3.6-27B \citep{qwen36_27b} generates a short utterance that could plausibly occur immediately before the target speech.
The prompt includes a dataset-dependent conversational context and asks the model to prefer a natural question without copying the target response.
For the English data used in our experiments, the generated utterance is limited to 18 words.
Invalid, empty, multi-line, overly long, or copied generations are rejected and regenerated.

\paragraph{User-speech synthesis.}
We synthesize the generated utterance with Qwen3-TTS-12Hz-1.7B-CustomVoice \citep{hu2026qwen3}.
One of two English speaker presets is selected deterministically from the source-speaker identity, so clips from the same source speaker use a consistent user voice.
The synthesis instruction requests a brief, natural conversational style at a moderate speaking rate.
During data loading, the waveform is converted to mono, resampled to $16$~kHz, and partitioned into 50 audio chunks of $0.4$~s each.

\paragraph{Response-time placement.}
The generated user utterance is placed at the beginning of the 20-s episode.
We deterministically sample a post-utterance thinking delay $d_{\mathrm{think}}
\in \{0.4, 0.8, 1.2\}$ s and align the beginning of the target response to the next chunk boundary:
\begin{equation}
t_{\mathrm{target}} = \Delta
\left\lceil
\frac{d_{\mathrm{user}} + d_{\mathrm{think}}}{\Delta}
\right\rceil, \quad \Delta = 0.4\,\mathrm{s},
\end{equation}
where $d_{\mathrm{user}}$ is the duration of the synthesized user speech.
This guarantees that target speech and motion do not begin before the user has finished speaking.
The target word timestamps and the retargeted co-speech motion are shifted by the same $t_{\mathrm{target}}$.

\paragraph{Listening and speaking motion composition.}
Before $t_{\mathrm{target}}$, the robot follows a fixed listening-motion sequence.
We use the first 9~s of a selected S1 listening sequence and loop it as needed.
A $0.4$-s linear blend is applied at loop boundaries.
At the target-response boundary, we blend from the listening trajectory into the retargeted co-speech motion over $0.4$~s.
When the target motion ends before the 20-s episode boundary, we similarly blend back to the listening motion.
The final listening padding is retained to keep every episode the same length but is excluded from the motion loss and motion-normalization statistics, since it is not part of the observed target motion.

\paragraph{Filtering and statistics.}
A source clip is retained only if its target audio, word-timestamped transcript, retargeted S1 motion, generated user waveform, and response onset all fit within the 20-s episode.
Starting from 29,152 English candidates, 110 clips are rejected because they do not contain valid timed words, and five are rejected because the target speech begins outside the episode.
The resulting Co-Speech set contains 29,037 episodes, as summarized in Table~\ref{tab:cospeech_stats}.

\subsection{Emotion Annotation for Second-Stage Training}
\label{app:emotion_annotation}

\begin{wraptable}{r}{0.4\textwidth}
    \centering
    \caption{Distribution of semantic emotion annotations.}
    \setlength{\tabcolsep}{10pt}
    \label{tab:emotion_stats}
    \begin{tabular}{lr}
        \toprule
        Emotion & Number of episodes \\
        \midrule
        \texttt{happy}    & 2,561 \\
        \texttt{sad}      & 1,402 \\
        \texttt{angry}    & 1,172 \\
        \texttt{surprise} & 731 \\
        \texttt{normal}   & 23,103 \\
        \texttt{urgent}   & 68 \\
        \midrule
        Total & 29,037 \\
        \bottomrule
    \end{tabular}
\end{wraptable}

For the second training stage, we annotate the semantic emotion of every target response using Qwen3.6-27B \citep{qwen36_27b}.
The model assigns exactly one of six labels:
\begin{equation}
\begin{aligned}
    m_t \in \{&
    \texttt{happy}, \texttt{sad}, \texttt{angry},\\
    &\texttt{surprise}, \texttt{normal}, \texttt{urgent}
    \}.
\end{aligned}
\end{equation}
In particular, fear, anxiety, and nervousness are assigned to \texttt{sad}, whereas \texttt{urgent} is reserved for direct warnings, requests for immediate action, or emergencies.
Neutral descriptions of dangerous events remain \texttt{normal}.
We use deterministic decoding and store both the predicted label and its confidence score.
All 29,037 Co-Speech episodes receive a valid annotation, with the distribution reported in Table~\ref{tab:emotion_stats}.
Each speaking motion chunk inherits the semantic emotion label $m_t$ of its target response, which is used directly for vocal-token construction and FiLM modulation.

Although the raw emotion distribution is highly imbalanced, we use all 29,037 annotated episodes as the sampling pool and balance training samples across the six emotion categories.
In each epoch, we draw the same number of episodes from every category.
With a total budget of $M=3{,}072$ sampled episodes per epoch, the sampling probability and quota for each emotion label are
\begin{equation}
\begin{aligned}
    \Pr(m_t) &= \frac{1}{6},\\
    n_{m_t} &= \frac{M}{6} = 512.
\end{aligned}
\end{equation}
Classes with more available episodes than their quota are randomly undersampled without replacement, whereas rare classes are oversampled through independently shuffled passes over their available episodes.
The resulting 3,072 samples are reshuffled at every epoch and then partitioned across distributed training workers.
This procedure prevents the dominant \texttt{normal} class from overwhelming the training objective while ensuring that all six emotion categories contribute equally to second-stage training. For the emotion-aware motion evaluation in Table~\ref{tab:interaction_results}, we construct 20 test cases with an approximately balanced distribution across the six semantic emotion categories: happy, sad, angry, surprise, normal, and urgent.

To assess the reliability of the generated emotion labels, we manually inspected 120 randomly sampled Co-Speech episodes. Two independent annotators evaluated (i) the consistency between the predicted emotion labels and the semantic content of the target responses, and (ii) the consistency between these labels and the corresponding motion sequences. The semantic consistency rate was 95.8\%, while the motion consistency rate was 95.0\%. These results support the reliability of the semantic emotion annotations and their correspondence with the underlying motion data.

\subsection{Test Set Construction and Data Independence}

\paragraph{Evaluation set construction.}

For the quantitative motion evaluation in Table~\ref{tab:quantitative_results}, we select
20 episodes of 20 seconds each from EI-Data. The evaluation set is constructed to cover representative interaction behaviors, including speaking, listening, and state transitions. Specifically, we consider three criteria: (1) clear distinctions between speaking and listening states, (2) diverse motion patterns, and (3) consistency between the expressed emotion and the corresponding motion. These criteria ensure that the evaluation includes meaningful interaction dynamics rather than segments dominated by static or repetitive behaviors.

\paragraph{Conditioning protocol for Table~\ref{tab:quantitative_results}.}
During motion training and the teacher-forced evaluation
in Table~\ref{tab:quantitative_results}, target-text tokens are assigned to their physical
intervals.
At decision step $t$, the motion expert can access target-text
history only through $I_t$, while the motion target lies
in $I_{t+1}$.
Target-text tokens from $I_{t+1}$ and later intervals are masked.
Online execution uses the same text-visibility cutoff,
with generated, playback-confirmed text replacing the
reference text history.
Reference speaking-state and emotion annotations provide
the vocal conditions during training and evaluation;
online execution obtains these conditions from the response
plan and playback schedule.

\paragraph{Data split and independence.}
To avoid potential overlap between training and evaluation,
we partition the data at the original recording or interaction
level. For the Seamless Interaction dataset, all segments
originating from the same dyadic interaction are assigned
exclusively to either the training or evaluation split,
including cases where the target and partner roles are
exchanged. For the Co-Speech datasets, segments derived
from the same source recording are similarly restricted
to a single split. This ensures that the training and
evaluation sets are disjoint at the original recording
or interaction level.

\paragraph{Robustness to test-set selection.}
To assess the influence of evaluation-set curation, we additionally evaluate EIP on 20 randomly sampled 20-second episodes from held-out recordings, without applying the interaction-diversity and emotion-consistency criteria used for the curated test set. All methods follow the same evaluation protocol as in Table~\ref{tab:quantitative_results}, using FGD, BC, and Div as evaluation metrics.
As shown in Table~\ref{tab:random_test_robustness}, EIP maintains its advantage over the evaluated baselines on the randomly sampled test set. Specifically, EIP achieves an FGD of 53.07, compared with 86.42 for MIBURI and 109.31 for RoboGesture, while also achieving better BC and diversity scores. Although FGD increases compared with the curated test set, the relative performance trend remains consistent. These results suggest that the improvements of EIP are not solely attributable to test-set curation, further supporting the robustness of the proposed method across different evaluation samples.

\begin{table}[t]
\centering
\setlength{\tabcolsep}{40pt}
\caption{Quantitative motion evaluation on randomly sampled episodes.
All methods are evaluated on 20 randomly sampled 20-second episodes
from held-out recordings, without applying the interaction-diversity
and emotion-consistency selection criteria used for the curated test set.}
\label{tab:random_test_robustness}
\begin{tabular}{lccc}
\toprule
Method & FGD$\downarrow$ & BC$\rightarrow$ & Div$\uparrow$ \\
\midrule
GT                    & --     & 0.754 & 1.089 \\
MIBURI$^*$             & 86.42  & 0.668 & 0.976 \\
RoboGesture$^*$        & 109.31 & 0.622 & 0.938 \\
\textbf{Ours (Final)}  & \textbf{53.07} & \textbf{0.719} & \textbf{1.004} \\
\bottomrule
\end{tabular}
\end{table}

\section{Incremental Generation and Playback Synchronization}
\label{app:token_commitment}

Algorithm~\ref{alg:incremental_streaming} describes response execution
after the interaction expert decides to respond. Let $t_0$ denote the
interaction step at which the response begins. During inference, the
interaction context $h_t$ is maintained as a committed KV cache
synchronized with speech playback. We denote the independent
response-decoding cache by $h^{\mathrm{dec}}$. Both caches use the same
frozen model parameters. The decoding cache is initialized from the
committed context $h_{t_0}$ and seeded with the response-start token.

\paragraph{Temporal convention.}
Let $I_t=[t\Delta,(t+1)\Delta)$, where
$\Delta=400\,\mathrm{ms}$.
Decision step $t$ follows $I_t$ and produces
speech and motion for $I_{t+1}$.
Thus, the subscript of $\hat{a}_t$ denotes its prediction step
rather than its playback interval.
This convention agrees with Eq.~\eqref{eq:state_action_alignment}, where the state block
covers $I_t$ and the target motion covers $I_{t+1}$.
Table~\ref{tab:temporal_scope} summarizes the temporal scope
of the conditioning information and outputs.

\begin{table}[t]
    \centering
    \caption{
        Temporal scope of conditioning information and outputs
        at decision step $t$.
        Decision indices and physical playback intervals
        are distinguished explicitly.
    }
    \label{tab:temporal_scope}
    \small
    \setlength{\tabcolsep}{4pt}
    \renewcommand{\arraystretch}{1.15}
    \begin{tabular}{p{0.25\linewidth}p{0.66\linewidth}}
        \hline
        \textbf{Quantity} &
        \textbf{Temporal scope at decision $t$} \\
        \hline

        Human audio $x_t$ &
        Audio received during the completed interval $I_t$. \\

        Action-visible context $h_t$ &
        Available interaction history, including target text
        committed through $I_t$. \\

        Robot state $q_t$ &
        State block associated with $I_t$. \\

        Vocal condition $v_t$ &
        Speaking state and response emotion for the upcoming
        output interval. \\

        Predicted motion $\hat{a}_t$ &
        Motion for $I_{t+1}$. \\

        Output speech $\hat{s}_t$ &
        Speech scheduled together with $\hat{a}_t$
        for $I_{t+1}$. \\

        Text aligned with $\hat{s}_t$ &
        Committed at the next decision, after its playback
        interval. \\

        \hline
    \end{tabular}
\end{table}

\paragraph{Coordination through shared interaction representations.}
EIP learns to predict the next motion interval from the
available interaction context, rather than requiring the text
spoken within that interval as an input.
The motion expert accesses the interaction model's layer-wise
representations, which also support response generation.
These representations provide a common semantic basis for
verbal and non-verbal behavior, allowing motion prediction
to learn associations between the preceding conversational
context and the accompanying future motion.
The vocal condition additionally conveys the speaking state
and semantic emotion of the planned response.
During online execution, response decoding advances in
$h_{\mathrm{dec}}$, while the action-visible context $h_t$
advances with playback.
The two processes therefore share an interaction-model basis
without requiring their decoding states to remain identical.

The buffer $\mathcal{B}$ stores synthesized audio, aligned text units,
and a playback cursor. Each synthesized phrase is appended to this
buffer, with its local timestamps shifted by the cumulative duration
of preceding phrases. Let $u_i$ be an aligned text unit and $\tau_i$
its start time on the response-relative audio timeline. The tokens
associated with playback interval $r$ are
\begin{equation}
\begin{gathered}
    \bar{y}_r =
    \operatorname{Tokenize}\!\left(
        \operatorname{Concat}
        \left\{u_i \mid r\Delta \leq \tau_i < (r+1)\Delta\right\}
    \right),\\
    \Delta = 400\,\mathrm{ms}.
\end{gathered}
\label{eq:playback_token_alignment}
\end{equation}

where text units are concatenated in temporal order. $r$ indexes
response-relative playback intervals, while $t$ indexes interaction
steps.

The first phrase is decoded and synthesized synchronously; subsequent
phrases are processed by a background worker. The response-level
emotion tag $m_t$ is obtained during initial decoding and retained across
phrases. The flag $\mathrm{done}$ becomes true only after decoding has
ended and the final synthesized phrase has been appended.

At decision step $t$, the outer loop supplies the available
human-audio update, the robot state $q_t$, and the interruption
flag $\iota_t$.
If the response remains active, the text aligned with the
preceding speech output, played during $I_t$, is committed
to $h_t$.
The motion expert then predicts $\hat{a}_t$ for $I_{t+1}$.
The speech chunk $\hat{s}_t$ is retrieved from the playback
buffer and dispatched for the same interval.
Its aligned text is committed at the next decision.
This ordering preserves the next-interval prediction
convention in motion training.
If insufficient audio is available, a silent chunk is emitted,
and the response playback and text cursors remain unchanged.

\begin{algorithm*}[!t]
\caption{Incremental Response Generation and Playback Synchronization}
\label{alg:incremental_streaming}
\footnotesize
\begin{algorithmic}[1]
\Require Frozen interaction model $M$, TTS model $S$,
committed interaction context $h_{t_0}$ stored as a KV cache
\Require Motion expert and chunk duration $\Delta=400\,\mathrm{ms}$
\Ensure Synchronized speech--motion output and updated interaction context

\State $h^{\mathrm{dec}}
    \gets \Call{ForkForResponse}{M,h_{t_0}}$
\State $\mathcal{B}\gets\varnothing$, $m_t\gets\bot$
\State $\mathrm{done}\gets\mathrm{false}$,
    $\mathrm{cancelled}\gets\mathrm{false}$,
    $\mathrm{sent}\gets\mathrm{false}$

\State \Call{ProducePhrase}{}
    \Comment{Decode and synthesize the first phrase}
\State \Call{LaunchAsync}{\textsc{ContinueResponse}}

\For{each interaction step $t \ge t_0$}
    \Statex \hspace{\algorithmicindent}
    \textit{The outer loop supplies updated $h_t$, $q_t$, and $\iota_t$.}

    \If{$\iota_t$}
        \State $\mathrm{cancelled}\gets\mathrm{true}$
        \State \Call{Clear}{$\mathcal{B}$}
        \State $h_t\gets\Call{CloseResponse}{M,h_t}$
        \State \Return $h_t$
    \EndIf

    \If{$\mathrm{sent}$}
        \State $F\gets\Call{AdvancePlayback}{\mathcal{B},\Delta}$
        \State $h_t\gets\Call{CommitFragment}{M,h_t,F,m_t}$
        \Comment{Commit text from completed interval $I_t$}
        \State $\mathrm{sent}\gets\mathrm{false}$
    \EndIf

    \If{$\mathrm{done}$ \textbf{and}
        \Call{Exhausted}{$\mathcal{B}$}}
        \State $h_t\gets\Call{CloseResponse}{M,h_t}$
        \State \Return $h_t$
    \EndIf

    \State $v_t
        \gets\Call{VocalCondition}{\mathrm{active},m_t}$
    \State $\hat{a}_t
        \gets\Call{GenerateMotion}{h_t,q_t,v_t}$
        \Comment{Predict motion for $I_{t+1}$}
    \State $\hat{s}_t
        \gets\Call{PeekAudio}{\mathcal{B},\Delta,\mathrm{done}}$

    \If{$\hat{s}_t=\bot$}
        \State $\hat{s}_t\gets\mathbf{0}$
            \Comment{Keep the response cursor unchanged}
    \Else
        \State $\mathrm{sent}\gets\mathrm{true}$
    \EndIf

    \State \Call{DispatchTogether}{$\hat{s}_t,\hat{a}_t$}
    \Comment{Jointly schedule both outputs for $I_{t+1}$}
\EndFor

\Statex
\Procedure{ProducePhrase}{}
    \State $(p,h^{\mathrm{dec}},m_t,f)
        \gets\Call{DecodePhrase}{M,h^{\mathrm{dec}},m_t}$
    \If{$p\neq\varnothing$}
        \State $(w,P)\gets\Call{SynthesizeAndAlign}{S,p}$
        \State \Call{AppendIfActive}
            {$\mathcal{B},w,P,\mathrm{cancelled}$}
    \EndIf
    \If{\textbf{not} $\mathrm{cancelled}$}
        \State $\mathrm{done}\gets f$
    \EndIf
\EndProcedure

\Statex
\Procedure{ContinueResponse}{}
    \While{\textbf{not} $\mathrm{done}$
        \textbf{and not} $\mathrm{cancelled}$}
        \State \Call{ProducePhrase}{}
    \EndWhile
\EndProcedure

\end{algorithmic}
\end{algorithm*}

\noindent
In Algorithm~\ref{alg:incremental_streaming},
\textsc{DecodePhrase} continues decoding until a valid punctuation
boundary or the response-end token.
\textsc{AppendIfActive} checks whether the response is active,
shifts phrase-local timestamps to the shared audio timeline,
and appends synthesized audio and aligned text to $\mathcal{B}$.
Cancellation checks and buffer updates are performed atomically.

\textsc{PeekAudio} retrieves the next available audio chunk without
advancing the playback cursor. It returns $\bot$ if insufficient audio
is available within the allowed wait; only the final incomplete chunk
is zero-padded.
\textsc{AdvancePlayback} advances past the previously emitted chunk
and retrieves its aligned text and boundary flags.
\textsc{CommitFragment} commits this text to $h_t$, adding the
response-start and emotion tokens for the first fragment and the
response-end token for the final fragment.

\textsc{CommitSilence} records silence preceding response
playback.
On interruption, \textsc{CloseResponse} appends the response-end token
if the response has entered the committed context, or a silence-control
token otherwise. State is scoped to each response, preventing cancelled
generation from appending audio to subsequent response.

\section{Rationale for the Attention Mask}
\label{app:mask_rationale}

The attention mask reflects the conditional generation process of EIP: the available interaction context, robot state, and vocal condition guide the generation of the next motion chunk, while the noisy action tokens are jointly updated through flow matching. The ordering $[Q_t; v_t; A_t^\tau]$ separates the conditioning inputs from the variables being denoised.

\paragraph{Interaction-context isolation and temporal visibility.}
Interaction-context tokens retain autoregressive attention and do not attend to motion-expert tokens. This preserves the frozen interaction expert's information dependencies and prevents its representations from changing with sampled motion. The interaction context can therefore be computed and cached independently of action denoising. Each motion chunk accesses only the context available at its decision step, preventing attention to later information when multiple chunks are processed together during training.

\paragraph{Ordering of state, vocal, and action tokens.}
Within each chunk, $Q_t$ represents the robot's current physical state, $v_t$ specifies the robot's speaking state and the semantic emotion of its response, and $A_t^\tau$ represents the action trajectory being generated. Allowing $v_t$ to attend to $Q_t$ lets the vocal-condition representation incorporate the current physical state. Keeping the reverse connection masked leaves the state representation independent of the vocal-token pathway. Action tokens attend to both blocks, combining the starting state with the desired behavioral condition. Neither state nor vocal tokens attend to action tokens, so their representations remain independent of the changing noisy trajectory throughout flow integration. This ordering is an inductive bias that maintains a separation between fixed conditions and generated actions.

\paragraph{Bidirectional attention within blocks.}
Tokens within the state block encode observations available at the decision boundary and can attend to one another. Tokens within the action block represent a trajectory generated jointly through flow matching. Bidirectional attention allows each action token to incorporate information from the entire candidate trajectory and model dependencies among its frames. These tokens are denoising variables available during the current computation, rather than future observations, so their mutual visibility does not violate temporal causality.

\paragraph{Isolation between motion chunks.}
Direct attention between motion chunks is masked. Blocking future chunks prevents access to unavailable robot states and future motion information. Blocking past motion chunks is an architectural choice that avoids dependence on their intermediate denoising representations: temporal information is instead carried through the accumulated interaction context and the updated robot state. Given the same conditioning inputs, each chunk  has the same local attention dependencies whether processed alongside other chunks during training or generated individually during streaming inference.

These restrictions are applied at every layer, preventing future context or other motion chunks from influencing the current prediction through indirect attention paths. This argument assumes that the input features and conditioning signals are available at the decision boundary.

\section{Sensitivity to Stage-I Loss Weights}
\label{app:loss_weight_analysis}

We examine how the Stage-I loss weights affect motion
quality. We fix $\lambda_{\mathrm{spk}}=1$ and vary
$\lambda_{\mathrm{lis}}$ and $\lambda_{\mathrm{idle}}$ around
the default setting $(1,2,1)$.
Table~\ref{tab:loss_weight_user_study} summarizes ratings from
five participants on a seven-point Likert scale, covering
Naturalness, State Adaptation, and Listening Diversity.
Average is the unweighted mean of these three dimensions. The default setting achieves the highest average rating of
6.00 and the highest State Adaptation score of 6.40, while
matching the highest Naturalness score of 5.80.
Reducing $\lambda_{\mathrm{lis}}$ from 2 to 1 lowers all three
ratings, resulting in an average of 5.33.
Increasing it to 4 yields the highest Listening Diversity
score of 6.20, but State Adaptation decreases to 4.00 and
the average falls to 5.27.
These ratings suggest that emphasizing listening supervision
can favor listening diversity without necessarily improving
behavioral adaptation across interaction states.

Changing the idle weight also produces trade-offs.
With $\lambda_{\mathrm{idle}}=0.5$, Listening Diversity
increases to 6.00, whereas State Adaptation decreases to
5.00, giving an average of 5.60.
With $\lambda_{\mathrm{idle}}=2$, Listening Diversity drops
to 4.00 and the average decreases to 5.27.
Naturalness remains at 5.80 for both variants.
Among the configurations evaluated in this small observer
study, $(1,2,1)$ therefore provides the strongest aggregate
rating, balancing naturalness, state adaptation, and
listening diversity.

\begin{table}[t]
\centering
\small
\setlength{\tabcolsep}{16pt}
\caption{Observer-based evaluation of Stage-I loss weights.
Five participants rate each configuration on a seven-point
Likert scale.}
\label{tab:loss_weight_user_study}
\begin{tabular}{ccc|cccc}
\toprule
$\lambda_{\mathrm{spk}}$ &
$\lambda_{\mathrm{lis}}$ &
$\lambda_{\mathrm{idle}}$ &
Naturalness$\uparrow$ &
\shortstack{State\\Adaptation$\uparrow$} &
\shortstack{Listening\\Diversity$\uparrow$} &
Average$\uparrow$ \\
\midrule
1 & 1 & 1   & 5.60 & 5.40 & 5.00 & 5.33 \\
1 & 4 & 1   & 5.60 & 4.00 & 6.20 & 5.27 \\
1 & 2 & 0.5 & 5.80 & 5.00 & 6.00 & 5.60 \\
1 & 2 & 2   & 5.80 & 6.00 & 4.00 & 5.27 \\
\midrule

1 & 2 & 1 & 5.80 & 6.40 & 5.80 & \textbf{6.00} \\
\bottomrule
\end{tabular}
\end{table}

\section{Ablation of Stage-II Emotion Conditioning}
\label{sec:emotion_ablation}
We compare matched fine-tuning without explicit emotion
conditioning (A), embedding-only conditioning (B), and
Full Stage II, using Stage I as a reference.
Variant A disables both the additive emotion embedding and
FiLM, while B retains the emotion embedding in the vocal
token without FiLM modulation.
All Stage-II variants start from the same Stage-I checkpoint
and use identical training data, class-balanced sampling,
training budgets, and backbone fine-tuning settings.
Emotion-aware motion success and FGD follow the evaluation
protocols and test sets of
Tables~\ref{tab:quantitative_results}
and~\ref{tab:interaction_results}, respectively.

\begin{table}[t]
    \centering
    \small
    \setlength{\tabcolsep}{50pt}
    \caption{
        Ablation of Stage-II emotion conditioning.
        Success denotes emotion-aware motion success.
        All Stage-II variants use matched training budgets
        and class-balanced sampling.
    }
    \label{tab:stage2_ablation}
    \begin{tabular}{lcc}
        \hline
        Configuration & Success (\%)$\uparrow$
                      & FGD$\downarrow$ \\
        \hline
        Stage I                & 40.0 & 69.33 \\
        A: Matched fine-tuning & 40.0 & 55.20 \\
        B: Embedding only      & 75.0 & 49.70 \\
        Full Stage II          & \textbf{95.0}
                               & \textbf{43.83} \\
        \hline
    \end{tabular}
\end{table}

As shown in Table~\ref{tab:stage2_ablation}, continuing
fine-tuning under the Stage-II data and sampling recipe
without explicit emotion conditioning (A) reduces FGD
from 69.33 to 55.20, while emotion-aware motion success
remains at 40\%.
Thus, this continued fine-tuning improves the FGD score
but does not improve the observed emotion-success rate
on the evaluated samples.
The emotion-success gain of Full Stage II is therefore
not reproduced by the matched fine-tuning control.

Adding the emotion embedding to the vocal token (B)
increases success from 40\% to 75\% and further reduces
FGD to 49.70.
Since A and B share the training data, sampling strategy,
and backbone optimization settings, this comparison
supports the benefit of explicit emotion injection through
the vocal-token pathway.
Full Stage II further increases success from 75\% to 95\%
and reduces FGD from 49.70 to 43.83, supporting an
incremental benefit from FiLM when the emotion embedding
is already present.

Together, these results support gains from the additive
emotion embedding and additional FiLM modulation under
the fixed Stage-II training recipe.

\section{Additional Interaction Analysis}
\label{app:qualitative}

\paragraph{Additional Evaluation on SMPL-X.}
To further evaluate EIP beyond the robot-specific motion
representation, we conduct an additional quantitative
evaluation on the SMPL-X skeleton with a
165-dimensional motion representation.
We follow the evaluation protocol described in the main paper, using the
EI-Data test set and the same evaluation metrics:
FGD, BC, and Div.

As shown in Table~\ref{tab:quantitative_results_smplx},
EIP outperforms all baselines across the three
metrics, achieving an FGD of 5.85, a BC of 0.547, and
a diversity score of 0.487.
Compared with the strongest baseline in terms of FGD,
RoboGesture$^\ast$, EIP reduces FGD from 25.18 to 5.85.
These results complement the robot-space evaluation
in the main paper and demonstrate the effectiveness
of EIP on both robot-specific and human motion
representations.

\begin{table}[t]
    \centering
    \caption{
    Quantitative motion evaluation on the SMPL-X skeleton.
    For baseline methods, $^\ast$ denotes training on our
    EI-Data in the target skeleton; otherwise, generated
    motions are directly retargeted to the target skeleton.
    }
    \label{tab:quantitative_results_smplx}
    \small
    \setlength{\tabcolsep}{50pt}
    \begin{tabular}{@{}lccc@{}}
        \toprule
        System
        & FGD$\downarrow$
        & BC$\to$
        & Div$\uparrow$ \\
        \midrule
        GT
        & -- & 0.685 & 0.499 \\
        \midrule
        MIBURI~\citep{mughal2026miburi}
        & 25.25 & 0.487 & 0.397 \\

        EMAGE~\citep{liu2024emage}
        & 32.21 & 0.485 & 0.468 \\

        Photoreal~\citep{ng2024photoreal}
        & 30.76 & 0.463 & 0.469 \\

        RoboGesture~\citep{wang2026robogesture}
        & 39.67 & 0.356 & 0.447 \\

        MIBURI$^\ast$~\citep{mughal2026miburi}
        & 26.22 & 0.529 & 0.471 \\

        RoboGesture$^\ast$~\citep{wang2026robogesture}
        & 25.18 & 0.454 & 0.400 \\
        \midrule
        \textbf{Ours (Final)}
        & \textbf{5.85}
        & \textbf{0.547}
        & \textbf{0.487} \\
        \bottomrule
    \end{tabular}
\end{table}

\begin{figure*}[htp]
    \centering
    \includegraphics[width=\linewidth]{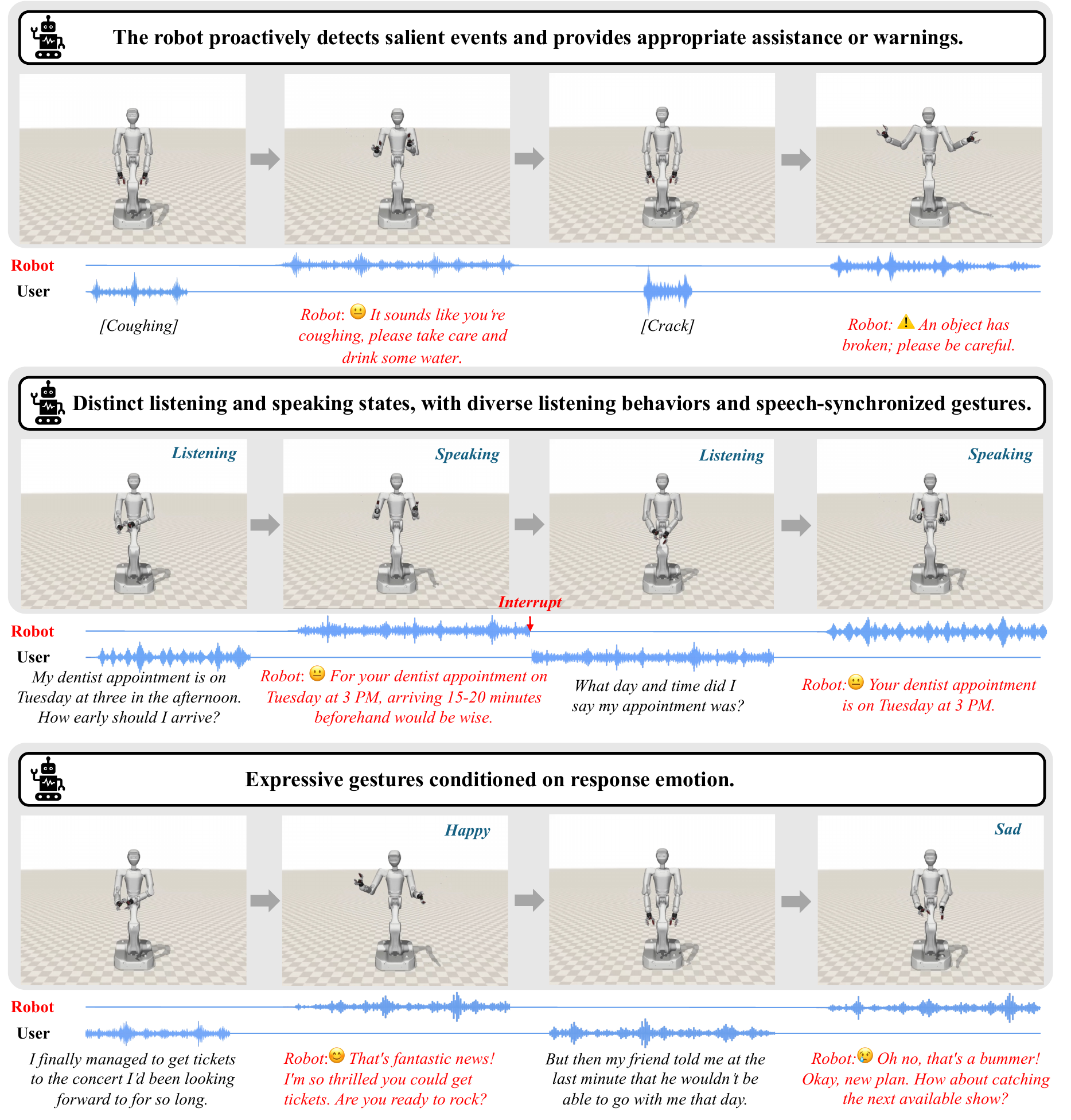}
    \caption{\textbf{Qualitative results of EIP in three representative interaction scenarios.} Top: proactive responses to salient acoustic events, such as coughing and object breaking. Middle: continuous turn-taking under user interruption, with distinct listening and speaking behaviors and speech-synchronized gestures. Bottom: emotion-aware gesture generation, where motion adapts to the response emotion.}
    \label{fig:simulation_result}
\end{figure*}

\paragraph{Qualitative Results.}
Figure~\ref{fig:simulation_result} shows additional EIP rollouts in simulation under the same three interaction scenarios evaluated in Section~\ref{sec:qualitative}.
In the top row, the robot produces proactive verbal and gestural responses to salient non-speech acoustic events, such as coughing and object breaking.
The middle row illustrates interruption-aware turn-taking: the robot switches from speaking to listening when the user barges in, then resumes with speech-synchronized gestures.
The bottom row shows emotion-aware gesture generation, where the generated gestures reflect the semantic emotion of the response.
These examples complement the physical-robot visualizations in the main paper.

\paragraph{Interruption response delay.}
We evaluate interruption response timing over 20 audio-conditioned rollouts.
For each interruption $k$, we measure the interval between acoustic onset
$t_k^{\mathrm{onset}}$ and the start of the first listening-conditioned
motion chunk $t_k^{\mathrm{listen}}$. Acoustic onsets are estimated from
normalized input audio using non-overlapping 10-ms RMS windows and a
threshold of 0.012. Pooling all interruption events across the 20 rollouts,
we compute
\begin{equation}
    \bar{L}=\frac{1}{N}\sum_{k=1}^{N}
    \left(t_k^{\mathrm{listen}}-t_k^{\mathrm{onset}}\right),
\end{equation}
where $N$ is the total number of interruption events.
The mean response delay is \textbf{220\,ms} on the output timeline and this delay measures the transition to listening-conditioned
motion on the output timeline.

\paragraph{Transition smoothness.}
On the same 20 rollouts, we quantify angular jerk for 20 upper-body joints
at 30\,Hz after six-frame $C^2$ stitching and 4-Hz offline zero-phase
low-pass filter. For each rollout $r$, we compute third-order
differences $\mathbf{j}_{r,i}=\Delta^3\mathbf{q}_{r,i}/h^3$, with
$h=1/30$\,s, before selecting transition windows. We report
\begin{equation}
    \bar{J}=\frac{1}{|\mathcal{W}|}
    \sum_{(r,i)\in\mathcal{W}}\lVert\mathbf{j}_{r,i}\rVert_2,
\end{equation}
where $\mathcal{W}$ is the union of available jerk samples whose stencil
midpoints lie within $\pm0.4$\,s of a speaking/listening transition across
all 20 rollouts. Windows are clipped at sequence boundaries without padding.
The pooled mean angular jerk norm is
$\bar{J}=\mathbf{239.03}\,\mathrm{rad}/\mathrm{s}^{3}$.
This metric characterizes acceleration variation during state transitions
in the final postprocessed trajectories.

\paragraph{Observer-based comparison with RoboGesture.}
15 participants rated
co-speech and interaction videos on a seven-point Likert scale
for Motion Naturalness (smooth, speech-synchronized body movements without noticeable stiffness, jitter, or abrupt transitions), Emotional Expressiveness (clear and appropriate expression of the response emotion through body movements), and Gesture Diversity (varied, non-repetitive gestures).
Since RoboGesture does not generate listening motions in our
inference setup, we exclude State Adaptation and Listening Diversity.
For proactive-event samples, we use EIP's generated spoken reply
as the speech input to RoboGesture for co-speech motion generation,
since RoboGesture cannot respond to these events directly.
For clips containing two user utterances, we provide only the first
utterance to both models, as RoboGesture does not support interruption
handling and produces only one response.
Participants evaluate body motion, with audio providing content
and emotion context. Ratings are aggregated across co-speech
and interaction samples.
These settings keep the comparison focused on perceived motion
quality while reducing the influence of differences in proactive
response and interruption capabilities.

\end{document}